\documentclass[letterpaper,12pt,titlepage,oneside,final]{book}

\newcommand{\href}[1]{#1} 

\usepackage{ifthen}
\newboolean{PrintVersion}
\setboolean{PrintVersion}{false} 

\usepackage{amsmath,amssymb,amstext} 
\usepackage[pdftex]{graphicx} 

\usepackage{framed}
\usepackage{xcolor}
\usepackage{mdframed}
\usepackage{enumerate}
\usepackage{graphicx,amssymb}
\usepackage{authblk}
\usepackage{amsmath}
\numberwithin{equation}{section}
\usepackage{titling}
\usepackage{caption}
\usepackage{graphicx}
\usepackage{float}
\restylefloat{figure}
\usepackage{lipsum}
\usepackage{algorithm}
\usepackage[noend]{algpseudocode}
\usepackage{multirow}
\usepackage[utf8]{inputenc}
\usepackage{booktabs}
\usepackage{siunitx}
\usepackage{rotating}
\usepackage{adjustbox}
\DeclareMathOperator*{\argmin}{argmin}
\usepackage{subcaption}
\usepackage{verbatim}

\usepackage[pdftex,letterpaper=true,pagebackref=false]{hyperref} 
\hypersetup{
    plainpages=false,       
    pdfpagelabels=true,     
    bookmarks=true,         
    unicode=false,          
    pdftoolbar=true,        
    pdfmenubar=true,        
    pdffitwindow=false,     
    pdfstartview={FitH},    
    pdftitle={uWaterloo\ LaTeX\ Thesis\ Template},    
    pdfnewwindow=true,      
    colorlinks=true,        
    linkcolor=blue,         
    citecolor=green,        
    filecolor=magenta,      
    urlcolor=cyan           
}
\ifthenelse{\boolean{PrintVersion}}{   
\hypersetup{	
    citecolor=black,%
    filecolor=black,%
    linkcolor=black,%
    urlcolor=black}
}{} 

\let\origdoublepage\cleardoublepage
\newcommand{\clearemptydoublepage}{%
  \clearpage{\pagestyle{empty}\origdoublepage}}
\let\cleardoublepage\clearemptydoublepage

\begin{document}


\pagestyle{empty}
\pagenumbering{roman}

\begin{titlepage}
        \begin{center}
        \vspace*{1.0cm}

        \Large
        {\bf Survey: Geometric Foundations of Data Reduction}

        \vspace*{1.0cm}
        
        Ce Ju \\
        \normalsize
        juce.sysu@gmail.com
       \vspace*{2.0cm}
       \end{center}
        \begin{center}\textbf{Abstract}\end{center}
This survey is written in summer, 2016. 
The purpose of this survey is to briefly introduce nonlinear dimensionality reduction (NLDR) in data reduction. 
The first two NLDR were respectively published in \emph{Science} in 2000 in which they solve the similar reduction problem of high-dimensional data endowed with the intrinsic nonlinear structure. 
The intrinsic nonlinear structure is always interpreted as a concept in manifolds from geometry and topology in theoretical mathematics by computer scientists and theoretical physicists. 
In 2001, the concept of Manifold Learning first appears as an NLDR method called Laplacian Eigenmaps. 
In a typical manifold learning setup, the data set, also called the observation set, is distributed on or near a low dimensional manifold $M$ embedded in $\mathbb{R}^D$, which yields that each observation has a $D$-dimensional representation. 
The goal of manifold learning is to reduce these observations as a compact lower-dimensional representation based on the geometric information. 
The reduction procedure is called the spectral manifold learning. 
In this paper, we derive each spectral manifold learning with the matrix and operator representation, and we then discuss the convergence behavior of each method in a geometric uniform language. 
Hence, the survey is named \emph{Geometric Foundations of Data Reduction}.

\end{titlepage}







\renewcommand\contentsname{Table of Contents}
\tableofcontents
\cleardoublepage
\phantomsection

\addcontentsline{toc}{chapter}{List of Tables}
\listoftables
\cleardoublepage
\phantomsection		



\pagenumbering{arabic}

\chapter{Introduction}
Data reduction dates back to 1901 when the English mathematician and biostatistician Karl Pearson published \emph{On lines and planes of closest fit to systems of point in space} \cite{peason1901lines}. In the paper, Pearson studied the data reduction problem and proposed the famous method of Principle Component Analysis (PCA). In 1958 and 1966, Torgerson and Gower proposed a similar method called the classical multidimensional scaling, also known as the Principle Coordinates Analysis \cite{torgerson1958theory}. The two early data reduction methods evolved during the decades and have many variants now. In 2000, two groups of scientists independently published their new methods ISOMAP \cite{tenenbaum2000global} and LLE \cite{roweis2000nonlinear} in \emph{Science}. People then began to realize the importance of data science. \\\\
Data reduction plays a role in data science. According to a famous scientific statement, the curse of dimensionality proposed by mathematician Bellman, most of the algorithms need exponentially more data samples in the high dimensional space to work efficiently as well as in the low one. One way to overcome this difficulty is to reduce the original data to a lower space. In particular, if the original data has too much meaningless information, data reduction becomes a necessary procedure before data analysis. For example, we can regard an image of a handwritten digit with $28\times 28$ pixels as a 784-dimensional vector. In general, the handwritten digit only occupies a little space. It means most of the coordinates of this high dimensional vector contains nothing. Thus, we can reduce these vectors to acquire a dense representation in the low dimension before data analysis without losing too much information. Another role for data reduction is that data scientists want to observe data before data analysis and then they reduce data to a visual space by the data reduction algorithms. \\\\
In the early age of data reduction, scientists only considered the linear case due to their simplicity. When ISOMAP and LLE came out in 2000, people began to pay attention to those data with nonlinear structure. This class of problem is called the manifold learning problem. In general, the framework of the manifold learning problem assumes the data are on (or near) a manifold without knowing their dimensions. The traditional methods such as PCA and MDS cannot be used as efficient ways in most cases, because the linear projection methods won't distinguish the nonlinear structure correctly. In chronological order, the main manifold learning algorithms include ISOMAP \cite[2000]{tenenbaum2000global}, LLE \cite[2000]{roweis2000nonlinear}, EigenMap \cite[2003]{belkin2003problems}, Hessian EigenMap \cite[2003]{donoho2003hessian}, MVU \cite[2004]{weinberger2006unsupervised}, LTSA \cite[2004]{zhang2004principal}, Diffusion Map \cite[2004]{lafon2004diffusion} and Vector Diffusion Map \cite[2011]{singer2012vector}. All the algorithms have their own characteristics. For example: ISOMAP is a method preserving the geodesic distance on the manifold; LLE considers the relation of nearby points as the linearity; EigenMap utilizes the graph Laplacian to approximate the Laplace-Beltrami operator; Hessian EigenMap is based on the fact that a function is linear if and only if it has everywhere vanishing Hessian; MVU keeps the metric isometry; LTSA aligns the bases of tangent space together; Lastly, Diffusion Map and Vector Diffusion Map are methods utilizing the diffusion process on the data graph. All these algorithms are formulated as an optimization problem based on their characteristics, and then scientists will apply spectral decomposition on some matrices to acquire the representation of data. Thus, these algorithms are also called the spectral manifold learning algorithms. Fortunately, there is a large amount of literature to solve this kind of optimization problems \cite{absil2009optimization}.\\\\

This survey is organized in the following way: In Section 2.1 of Chapter 2, we collect the properties from all the algorithms. Since every algorithm holds a special property, we relate them with the general geometric or topological properties and name them. In Section 2.2, we mainly talk about two methods of collecting the local information. One is called the method of the local linearity, which means the local neighbor points hold the linearity property. And, the other method is called the local PCA. It means the scientists do the PCA locally and collect an approximation of the tangent space. In Section 2.3, we introduce all the spectral manifold learning algorithms in chronological order. In this chapter, all the methods are presented in a language of matrix theory  for the computational mathematician and statistician. In Section 2.4, we compare the performance of each algorithms based on several factors.

In Chapter 3, we cover the convergence issue of the manifold learning algorithms, which is the core subject of this survey. We mainly analyze the convergence issue of the three algorithms EigenMap, Diffusion Map and Vector Diffusion Map. In Section 3.1, we discuss the relation between the geodesic distance and the embedding Euclidean distance. The distance or the function related to the distance parameter can be expanded with some curvature term and higher-order terms because of the intrinsic geometric structure of the manifolds embedded in the Euclidean space. In Section 3.2, we restate the three algorithms in the language of the operator theory on manifolds. These theoretical formulations are helpful for the convergence proof. In Section 3.3, we discuss the main idea for all the convergence issue of the three algorithms in a similar framework. The graph Laplacian converges to an averaging operator and the averaging operator converges to the Laplace-Beltrami operator. Finally, we write all the necessary math background in the Appendix including the basic matrix analysis, the linear manifold reduction algorithm, the Laplace-Beltrami operator and the Hessian tensor on manifolds, the heat operator on manifolds and the basic spectral graph theory.

\chapter{Manifold Learning Algorithms}
Suppose we have a connected compact smooth Riemannian manifold $(M, g)$ with dim$M=d$ and in addition a Riemannian immersion (or Riemannian embedding) $\iota: (M^d, g) \hookrightarrow (\mathbb{R}^D$, can). The metric $g$ for manifold $M$ is induced from the canonical metric on ($\mathbb{R}^D$, can). This means the metric locally performs like the Euclidean inner product on the tangent space of the manifolds. Suppose the data samples $\{x_1,\dots, x_N\}$ $(x_i\in\mathbb{R}^D)$ are distributed on or near some embedding manifold $\iota(M)$ with an unknown intrinsic dimension $d\ll D$. We will denote the point on $M$ by $x_i$ rather than $\iota(x_i)$ in most cases except for in the vector diffusion map algorithm. We call the space of data sample the \emph{sample} space and the space after reduction the \emph{feature} space. The aim of the reduction algorithm is to find a lower-dimensional representation of data in the feature space $\{y_1, \dots, y_N\}$ $(y_i \in \mathbb{R}^d)$ for the original data samples $\{x_1,\dots, x_N\}$ $(x_i\in\mathbb{R}^D)$. In the spectral method of data reduction, people could always write down a reduction function in an explicit way. In general, the manifold learning algorithms have two steps. In the first step, they extract the local information from the data samples obeying some specific properties. Then in the second step, they patch them to global information and obtain a lower-dimensional representation of the data.

\section{Properties of Algorithms}
Given two manifolds $(M, g_M)$ and $(N,g_N)$, if there exists a diffeomorphism $F: M\rightarrow N$ satisfying
\begin{align*}
       g_N(dF(v),dF(w))=g_M(v, w) \hspace*{1em} v, w \in TM,
\end{align*}
we call $F$ a (Riemannian) isometry. We always assume the embedding map $\iota: M^d \hookrightarrow \mathbb{R}^D$ is isometric.\\\\
For a pair of points $x_i$ and $x_j$ on $M$, we define the path space 
\begin{align*}
           \Omega_{ij} :=\{\gamma: [0,1]\rightarrow M| \gamma \in C^\infty([0,1]) \text{ and } \gamma(0)=x_i, \gamma(1)=x_j \}.
\end{align*}
Then the distance on Riemannian Manifolds $(M,g)$ between $x_i$ and $x_j$ is
\begin{align*}
        d_M(x_i, x_j):=\inf_{\gamma \in \Omega_{ij}} \int_{[0,1]} g(\dot{\gamma(t)},\dot{\gamma(t)})^\frac{1}{2} dt.
\end{align*}\\
The following parts are the two groups of properties that the algorithms obey:

\subsection{Geometric Properties}
\begin{itemize}
\item \textbf{Global Geodesic Isometry}: There exists a coordinate representation $\{y_1, \dots, y_N\}$ $(y_i \in \mathbb{R}^d)$ in the feature space satisfying 
\begin{align*}
               d_M(x_i, x_j) = ||y_i-y_j||_{\mathbb{R}^d}.
\end{align*}

\item \textbf{Local Geodesic Isometry}: For any point $x \in M$, there is a small neighborhood $U_x\subset M$ such that the distance between $x$ and any other point $y\in U_x$ on the manifold is equal to the Euclidean distance between their corresponding points in the feature space.

\item \textbf{Metric Isometry}: Given two metric spaces $(X,d_X)$ and $(Y, d_Y)$, there exists a map $F: X\rightarrow Y$ satisfying
\begin{align*}
d_Y(F(x_1), F(x_2))=d_X(x_1, x_2)   \hspace*{1em} x_1, x_2 \in X.
\end{align*}
\end{itemize}
\subsection{Topological Properties}
\begin{itemize}
\item \textbf{Local Topology Information}: The algorithm preserves the local orientation or the angle between points. 
\item \textbf{Connectedness in the Feature Space}: The algorithm requires the feature space to be open and connected in $\mathbb{R}^d$.
\end{itemize}

\section{Extracting the Local Information}
In the section, we only talk about two methods of extracting the local information of the data manifolds. The purpose for extracting the local information is mainly because we need to hold similar structure in the low dimensional representation based on these information. 

\subsection{Method A: (Local Linearity)}
This method views any data point as a linear combination of its neighboring $K$ points. Firstly, we need to compute the weights of the linear combination. We applies the optimization method to get the weights by minimizing the following total cost function, for all $i=1,\dots,N$ and $x_j \in N(x_i)$
\begin{align*}
        \mathcal{E}(W)&=\sum_i |x_i-\sum_{x_j} W_{ij}\cdot x_j|^2,  \\         
        \sum_{x_j} W_{ij}&=1,
\end{align*}
where the set $N(x_i)$ is the $K$-neighbor point set of $x_i$ and $W_{ij}$ are non-symmetric weights. \\\\
Fix an index $i$. We find the weights $W_{ij}$ in group $N(x_i)$ independently of the weights in other groups. Thus, we have only to minimize the cost for one group $N(x_i)$ and sum them up:
\begin{align*}
      \mathcal{E}_i(W)&=|x_i-\sum_{x_j} W_{ij}\cdot x_j|^2,\\
           \sum_{x_j} W_{ij}&=1.
\end{align*}
By the method of Lagrange multiplier, we get an explicit solution
\begin{align*}
       W_{ij}=\frac{\sum_k C_{jk}^{-1}}{\sum_{l,m} C_{lm}^{-1}},
\end{align*} 
where $C_{jk}:=(x_i-x_j)\cdot(x_i-x_k)^T$.
\subsection{Method B: (Local Principle Component Analysis)}
The main idea of this algorithm is to apply the PCA locally as its name implies. The classical Principle Component Analysis (PCA) algorithm is described in Section B.1 of Appendix B. The first step of the local PCA is to choose a scale parameter $\epsilon$. Then for each point, we define a neighborhood of $x_i$ on the manifold within Euclidean distance $\epsilon$
\begin{align*}
        B_{\epsilon}(x_i):=\{x_j: 0< ||x_j-x_i||<\epsilon\}.
\end{align*}
One issue for local PCA is how to pick the scale parameter. We need choose it such that $d \leq |B_{\epsilon}(x_i)|\ll D$. If we pick $\epsilon = O(n^{-\frac{2}{d+1}})$, then $|B_{\epsilon}(x_i)| =O(n^{\frac{1}{d+1}})$ \cite{singer2012vector}. After translating the points in $B_{\epsilon}(x_i)$ by $x_i$, we reduce $B_{\epsilon}(x_i)-x_i$ by PCA method. Since the points in $B_{\epsilon}(x_i)$ is distributed on (or near) $T_{x_i}M$, the estimated reduction dimension of $B_{\epsilon}(x_i)-x_i$ is close to the true dimension of $M$. Since the noise in the samples, scientists prefer to set a global dimension as the mean or the median of the local dimensions in order to 

\section{Patching to Global Information} 
In this section, we go over all the existing spectral manifold learning algorithms. The spectral manifold learning algorithm means the manifold learning algorithm applies a procedure of the eigenvalue decomposition in the algorithm. We will present the methods in chronological order. The reference includes ISOMAP \cite{tenenbaum2000global, bernstein2000graph}, LLE \cite{roweis2000nonlinear}, EigenMap \cite{belkin2003problems, belkin2003laplacian, von2008consistency}, Hessian EigenMap \cite{donoho2003hessian}, MVU \cite{weinberger2006unsupervised}, LTSA \cite{zhang2004principal}, Diffusion Map \cite{lafon2004diffusion, coifman2006diffusion} and Vector Diffusion Map \cite{singer2012vector}.

\subsection{ISOMAP (2000, J.B. Tenenbaum, V.D. Silva and J.C. Langford)}
This algorithm obeys the global geodesic isometry property. The main idea of ISOMAP algorithm is to implement the classical Multidimensional Scaling algorithm on the data samples with the dissimilarities by summing up a sequence of the path length on the weighted graph. The path length is chosen as the geodesic distance between nearby points on the manifold. The convergence issue shows that for a small $\epsilon$ ($\epsilon$-neighbor algorithm) or a suitable $K$ (KNN algorithm), the distance between points in feature space is very close to the geodesic distance between points in the data space \cite[Main Theorem A B \& C]{tenenbaum2000global}. The scheme for this algorithm is as follow,

\begin{algorithmic}[1]
\State $\text{Construct a neighbor graph by $\epsilon$-neighbor or KNN algorithm}$
\State $\text{Compute the shortest paths between nodes by Floyd-Warshall algorithm}$   
\State $\text{Construct the embedding by the Multidimensional Scaling method}$ 
\end{algorithmic}
\noindent We derive the new representation of data by multidimensional scaling. Since the graph distance $d_G(i, j)$ is collected by Floyd-Warshall algorithm, then the Gram matrix $G_g :=C_N\cdot (-\frac{1}{2}d_G^2(i, j))\cdot C_N$, where $C_N$ is the centering matrix (See B.2 on Appendix B). Then we decompose the Gram matrix $G_g$ by the spectral decomposition, i.e. 
\begin{align*}
               G_g = V\cdot \Lambda \cdot V^T
\end{align*}
where $V=[v_1,\dots, v_n]$ and $\lambda=$diag$(\lambda_1,\dots,\lambda_n)$. Then the reduction coordinates are given by $Y=(\sqrt{\lambda_1}\cdot v_1,\dots,\sqrt{\lambda_d}\cdot v_d)^T$. For more details of multidimensional scaling, please refer to B.2 in Appendix B.\\\\
 In particular, we have $y_i=(\sqrt{\lambda_1}\cdot v_1(i),\dots,\sqrt{\lambda_d}\cdot v_d(i))$.
\subsection{LLE (2000, L.K. Saul and S.T. Roweis)}
This algorithm collects the local information by the Method A (Local Linearity) in Section 2.2.1. After computing the weights between nearby points. Saul and Roweis recover the points preserving the same weights in the feature space. The numerical method for this recovering is by minimizing the embedding cost function 
\begin{align}
        \Phi(Y)=\sum_{i=1}^N |y_i-\sum_jW_{ij}y_j|^2.
\end{align}
To avoid degenerate solutions, we add two constraints. We require that the new coordinators $\{y_1,\dots,y_N\}$ are centering on the origin and have unit covariance, i.e.
\begin{align*}
       \sum_{i=1}^N y_i&=0,\\
       \frac{1}{N}\sum_{i=1}^N y_i\cdot y_i^T &=I_d.
\end{align*}
Define the $N\times N$ matrix $M$ be $(I-W)^T(I-W)$ and $M$ can be written as 
\begin{align*}
M_{ij}:=\delta_{ij}-W_{ij}-W_{ji}+\sum_k W_{ki}W_{kj}.
\end{align*}
We can simplify $\Phi(Y)$ as $tr(YMY^T)$ and then the objective function (2.3.1) becomes an optimization problem with orthogonality constraints, i.e. 
\begin{align*}
        \min_{\frac{1}{\sqrt{N}}Y^T\in \mathcal{V}_d(\mathbb{R}^N)} & tr[Y(I-W)^T(I-W)Y^T],\\
        s.t. \hspace*{1em}  & \sum_{i=1}^N y_i=0,
\end{align*}
where we write $Y=(y_1, \dots, y_N)$.\\\\
For the optimization problem with orthogonality constraints, we have a routine argument for the solution. See Section A.3 in Appendix A. We apply the SVD method on $M$ and get $M=V\cdot \Sigma \cdot V^T$. After truncating the smallest eigenvector $v_0\approx 0$, the truncated $Y^T$ consist of the eigenvectors $\{v_1,\dots, v_d\}$ corresponding to the 2\textsuperscript{nd} to the $(d+1)$\textsuperscript{st} smallest eigenvalues of $M$. In particular, we have $y_i=(v_1(i),\dots, v_{d}(i))$.
\subsection{EigenMap (2003, M. Belkin and P. Niyogi)}
This algorithm utilizes the graph Laplacian on the data graph. For the basic spectral graph theory, please see Appendix E for details. Firstly, we regard the data as the node on a graph and connect the nodes by $\epsilon$-Neighborhood algorithm or KNN algorithm. Secondly, we define the graph weights between data by the Gaussian kernel, i.e.
\begin{align*}
   w_{ij}:= e^{-\frac{||x_i-x_j||^2}{t}}.
\end{align*}
Then we get a weighted graph $W$ and we construct the Laplacian matrix on $W$ by
\begin{align*}
        \mathcal{L}:=D-W.
\end{align*}
The aim of EigenMap is to find a representation $Y=[y_1,\dots,y_N]^T$ $(y_i \in \mathbb{R}^D)$ minimizing the weight objective function
\begin{align}
             \sum_{i, j=1}^N ||y_i-y_j||_{\mathbb{R}^D} \cdot w_{ij}=tr(Y^TLY),
\end{align}
which has the orthogonal constraints $Y^TDY=I$.\\\\
The solution to (2.3.2) is the same to the following generalized eigenvector problem 
\begin{align*}
    \mathcal{L}f=\lambda D f,
\end{align*}
where the variable $f\in \mathbb{R}^N$.\\\\
Pick the smallest $d+1$ eigenvalues of $\mathcal{L}$ as $\lambda_0 (=0) \leq \lambda_1 \leq \dots \leq \lambda_d$. And, the corresponding eigenvectors are $f_0, f_1,\dots, f_d$. We truncate $f_0$ since it is a constant vector $c\cdot 1$. Then, the reduction map $\mathcal{E}(x_i)$ is given by
\begin{align*}
    \mathcal{E}(x_i):=(f_1(i), \dots, f_d(i)).
\end{align*}
\subsection{Hessian EigenMap (2003, D.L. Donoho and C. Grimes)}
This algorithm assumes the local geodesic isometry and the connectedness in the feature space. It utilizes the facts that a function is linear if and only if it has everywhere vanishing Hessian and the null space of $\mathcal{H}^{\text{iso}}$ is $(d+1)$-dimensional consisting of constant functions and isometric coordinates.\\\\
Consider two quadratic forms defined on a set of $C^2(M)$ functions
\begin{align*}
      \mathcal{H}^{\text{iso}}(f)&:=\int_M ||H_f^{\text{iso}}||_F^2,\\
      \mathcal{H}^{\text{tan}}(f)&:=\int_M ||H_f^{\text{tan}}||_F^2,
\end{align*}
where $H_f^{\text{tan}}$ in the norm is the Hessian operator in an orthonormal coordinates and $H_f^{\text{iso}}$ is the Hessian operator in an isometry coordinates. \\\\
In general, $\mathcal{H}^{\text{iso}}$ cannot be directly computed. Donoho and Grimes utilize $\mathcal{H}^{\text{tan}}$ to make it computable \cite[Theorem 1]{donoho2003hessian}, since the two Hessian operators obey  
\begin{align*}
\mathcal{H}^{\text{iso}}(f)= \mathcal{H}^{\text{tan}}(f)
\end{align*}
Hessian EigenMap collects the local information by Method B (Local PCA) in Section 2.2.1. Local PCA collects an orthogonal basis of the tangent space at each point. For each point $x_i$ in the data set, the neighbor points set $\{x_{i_1},\dots,x_{i_k}\}$ is denoted as $N(x_i)$. We reduce the translation of the neighbor points set  $M_i:=(x_{i_1}-\bar{x},\dots, x_{i_k}-\bar{x})$ by SVD method and get
\begin{align*}
       M_i=U \cdot \Sigma \cdot V^T
\end{align*}
where $U:=[u_1,\dots, u_D] \in \mathbb{R}^{D\times D}$, $V:=[v_1,\dots, v_k] \in \mathbb{R}^{k\times k}$ and a diagonal matrix $\Sigma \in \mathbb{R}^{D\times k}$.\\\\
Then the left eigenvector set $\{u_1,\dots, u_D\}$ is an orthogonal basis of $T_{x_i}M$ and the right eigenvector set $\{v_1,\dots, v_k\}$ consists of the tangent coordinate functions, where the tangent coordinate function means $v_i$ is a function with the value $[v_i(x_{i_1}),\dots, v_i(x_{i_k})]$ on $N(x_i)$.\\\\
Recall that the null space of $\mathcal{H}^{\text{tan}}$ is $(d+1)$-dimensional consisting of constant functions and isometric coordinates. Then we construct the local Hessian matrix $H_i(x_i)$ from $[v_i(x_{i_1}),\dots, v_i(x_{i_k})]$. Firstly, we choose the $d$ leading columns from $U$ and $V$. Then we compute the Hadamard product $Q_i:=(v_k \circ v_m)_{1\leq k\leq m\leq d}$. Finally, we orthonormalize the matrix $P_i:=[1, V_i, Q_i]$ and get the orthonormalized matrix 
\begin{align*}
\bar{P}_i=[1, V_i, \bar{Q}_i].
\end{align*}
Then the local Hessian matrix is given by $H_i= \bar{Q_i} \cdot \bar{Q_i}^T$.\\\\
Now we need to patch all the local information together. Firstly, we initialize the global Hessian $H$ to an $n\times n$ zero matrix. Then we update it by a submatrix kernel $H(N(i),N(i))$ in each time and obey the rule
\begin{align*}
H(N(x_i),N(x_i)):=H(N(x_i),N(x_i))+H_i,
\end{align*}
where $H(N(x_i),N(x_i))$ denote the submatrix of $H$ with rows and columns indexed by $N(x_i)$.\\\\
Minimize the $\mathcal{H}^{\text{tan}}$ over all the linear function $f$ in $C^2(M)$
\begin{align*}
             \min_f f^T  \mathcal{H}^{\text{tan}}  f,
\end{align*}     
where $f$ satisfies the orthogonal constraints.\\\\
Then the truncated $Y$ consists of the eigenvector $\{v_1,\dots, v_d\}$ corresponding to the 2\textsuperscript{nd} to the $(d+1)$\textsuperscript{st} smallest eigenvalues of the global Hessian $H$.\\\\
In particular, we have $y_i=(v_1(i),\dots, v_{d}(i))$.
\subsection{MVU (2004, K.Q. Weinberger and L.K. Saul)}
This algorithm is designed based on the metric isometry in Section 2.1.1. The metric isometry means the distance between any two points in the sample space $X$ is equal to the one in the feature space $Y$, i.e. 
\begin{align*}
                     ||y_i-y_j||_{\mathbb{R}^d}=||x_i-x_j||_{\mathbb{R}^D}.
\end{align*}
Suppose the Gram matrix $G_{ij}=\langle x_i, x_j \rangle_{\mathbb{R}^D}$ and the Gram matrix $K_{ij}=\langle x_i, x_j \rangle_{\mathbb{R}^d}$. Then the metric isometry is equivalent to 
\begin{align*}
         K_{ii}+K_{jj}-K_{ij}-K_{ji}=G_{ii}+G_{jj}-G_{ij}-G_{ji}.
\end{align*}
The objection function is to maximize the cost function 
\begin{align*}
        \mathcal{E}(Y):&=\frac{1}{2N}\sum_{i,j=1}^N |y_i-y_j|^2\\
                                &=\sum_{i=1}^N |y_i|^2 \\
                                &=\sum_{i=1}^N K_{ii} \\
                                &=tr(K).
\end{align*}
Centering $y_i$ to 0, we get a semidefinite programming problem
\begin{align*}
                     \max_{K\succeq 0}&\hspace*{0.5em}tr(K)\\
                                s.t. \hspace*{0.5em}  \sum_{i,j}K_{ij}&=0,\\
                          K_{ii}+K_{jj}-K_{ij}-K_{ji}&=G_{ii}+G_{jj}-G_{ij}-G_{ji}, \\
                                                                &\text{ for any connected $x_i$ and $x_j$.}               
\end{align*}
There exists a wealth of literature for solving SDPs efficiently. Suppose $K^*$ is the optimizer of the semidefinite programming problem. It is a Gram matrix by the constraint and thus we can decompose it by the spectral decomposition.\\\\
In particular, we have $y_i=(\sqrt{\lambda_1}\cdot v_1(i),\dots,\sqrt{\lambda_d}\cdot v_d(i))$.

\subsection{LTSA (2004, Z.-Y. Zhang and H.-Y. Zha)}
This algorithm collects the local information by Method B (Local PCA) in Section 2.2.1. Local PCA method collects an orthogonal basis of the tangent space at each point. For each point $x_i$ in the data set, the neighbor points set $\{x_{i_1},\dots,x_{i_k}\}$ is denoted as $N(x_i)$. We reduce the translation of the neighbor points set  $M_i:=(x_{i_1}-\bar{x},\dots, x_{i_k}-\bar{x})$ by SVD method and get 
\begin{align}
                    x_{i_j} =\bar{x_i} +V_i \beta_{i_j}+\xi_j^i \hspace*{1em} j\in\{1,\dots,k\},
\end{align}
where $\bar{x_i}$ is the mean of $N(x_i)$, $V_i$ is belong to $\mathcal{V}_d(\mathbb{R}^D)$, $\beta_{i_j}$ is the coordinates for $x_{i_j}$ on Stiefel manifold and $\xi_j^i$ is the reconstruction error written as $(I-V_i V_i^T)\cdot (x_{i_j}-\bar{x_i})$. \\\\
The global coordinates $\tau_i$ should be equal to the local coordinate coefficients $\beta_i$. Comparing with equation (2.3.3), we have 
\begin{align}
\tau_{i_j} =\bar{\tau_i} +L_i \beta_{i_j}+\epsilon_j^i \hspace*{1em} j=1,\dots,k
\end{align}
In a language of matrices, suppose the $d\times k$ global coordinate matrix $G_i=[\tau_{i_1},\dots, \tau_{i_k}]$ and the local reconstruction error matrix $E_i=[\epsilon_1^i,\dots,\epsilon_k^i]$. Then equation (2.3.4) can be written as
\begin{align*}
G_i=\frac{1}{k}G_i\cdot 1\cdot 1^T+L_i\beta_i+E_i
\end{align*}
To minimize the reconstruction error $E_i$, we consider 
\begin{align}
\sum_{i=1}^N ||E_i||^2=\sum_{i=1}^N ||G_i(I_d-\frac{1}{k}1\cdot 1^T)-L_i \beta_i||^2.
\end{align}
There are two variables $G_i$ and $L_i$ in (2.3.5). Firstly, we fix $G_i$ and solve for $L_i$. The explicit solution to $L_i$ and $E_i$ written in Moor-Penrose generalized inverse of $\beta_i$ are
\begin{align*}
L_i&=G_i(I_d-\frac{1}{k}1\cdot 1^T)\beta_i^+ = G_i \beta_i^+, \\
E_i&=G_i(I_d-\frac{1}{k}1\cdot 1^T)(I-\beta_i^+\beta_i).
\end{align*}
Define the $d\times N$ global coordinates matrix $G$ and the $N\times k$ selection matrices $S_i$ such that
\begin{align*}
      G\cdot S_i = G_i \hspace*{1em} i=\{1,\dots,N\}.
\end{align*}
Then we denote the global selection matrix $(S_1,\dots,S_N)$ as $S$ and the global reconstruction error matrix  $\text{diag}(W_1,\dots,W_N)$ as $W$, where $W_i=(I_d-\frac{1}{k}1\cdot 1^T)(I-\beta_i^+\beta_i)$. Finally, we need to determine $G_i$ by solving the following optimization problem.\\\\
Notice that 
\begin{align*}
       \sum_{i=1}^N ||E_i||_F^2=||GSW||_F^2.
\end{align*}
Then $G_i$ can be solved by the optimization problem with orthogonal constraints
\begin{align*}
         \max_{G^T\in \mathcal{V}_d(\mathbb{R}^N)} &\hspace*{0.5em}tr(G[SWW^TS^T]G).
\end{align*}
For the optimization problem with orthogonal constraints, please see Section A.3 in Appendix A.
\subsection{Diffusion Maps (2004, S. Lafon)}
Similar with EigenMap, we generate a graph $G=(V, E, w)$ by the data set. Diffusion Maps utilizes the probability that one walks from a node to the others randomly on a graph as the distance between a pair of points in the feature space. Let $P$ be the transition matrix for the data graph $G$ and $P^t$ as the $t$\textsuperscript{th} step transition matrix. We can write $P=D^{-1}W$, where the weight matrix $W$ has entry $w(u,w)$ and the diagonal matrix $D$ has diagonal entry $w(u)$. See Section E.2 of the Appendix E for the notations and conventions of the random walk on a graph. \\\\
Consider the symmetric matrix $S:=D^{-\frac{1}{2}}\cdot P\cdot D^{\frac{1}{2}}$. We decompose $S$ by the spectral decomposition  
\begin{align*}
                    S=V \cdot \Lambda \cdot V^T,
\end{align*}
where $V^T\cdot V=I_N$ and the diagonal matrix $\Lambda$ has a descending order diagonal entries $\lambda_1\geq \dots \geq \lambda_N$. \\\\
Note that
\begin{align*}
       P=(D^{-\frac{1}{2}}V)\cdot \Lambda \cdot (D^{\frac{1}{2}}V)^T.
\end{align*}
Then we reduce matrix $P$ by SVD, so that 
\begin{align*}
      P=\Phi \cdot \Lambda \cdot \Psi^T,
\end{align*}
which implies $\Phi:=[\phi_1,\dots,\phi_N]=D^{-\frac{1}{2}}V$ and $\Psi:=[\varphi_1,\dots,\varphi_N]=D^{\frac{1}{2}}V$.\\\\
We can write $P$ as the sum of rank 1 matrices
\begin{align*}
          P=\sum_{k=1}^N [\lambda_k\cdot \phi_k] \varphi_k^T,
\end{align*}
It follows that the $t$\textsuperscript{th} step transition matrix $P^t=\sum_{k=1}^N [\lambda_k^t\cdot \phi_k] \varphi_k^T$. And then, the diffusion map $\mathcal{D}_t$ is given as follows, 
\begin{align*}
 [\mathcal{D}_t](u):=(\lambda_1^t\cdot \phi_1(u), \dots, \lambda_n^t\cdot \phi_n(u)).
\end{align*}
\textbf{Note}
\begin{itemize}
      \item We can truncate the first coordinate $\lambda_1^t\cdot \phi_1$ because the $k$\textsuperscript{th} step transition matrix $P^k$ always has the biggest simple eigenvalue 1 and thus $\lambda_1^t\cdot \phi_1(u)=1$ for any node $u\in V$. The truncated diffusion map with $d$ dimensions is the $d$ leading coordinates of the diffusion map.
      \item The Diffusion map $[\mathcal{D}_t]$ has the inner product in the feature space
               \begin{align*}
                       \langle [\mathcal{D}_t](u),  [\mathcal{D}_t](v)\rangle_{\mathbb{R}^d}=\sum_{k=1}^n \left(\frac{P^t(u,s)}{\sqrt{w(s)}}\cdot \frac{P^t(v,s)}{\sqrt{w(s)}}\right).
               \end{align*}
     \item The diffusion distance $d_{\text{Diffusion, t}}^2(u,v)$ is given by
               \begin{align*}
                        d_{\text{Diffusion, t}}^2(u,v) =\sum_{s\in G} \frac{(P^t(u,s)-P^t(v,s))^2}{w(s)}.   
               \end{align*}
\end{itemize}
\subsection{Vector Diffusion Maps (2011, A. Singer and H.-T. Wu)}
This algorithm collects the local information by Method B (Local PCA) in Section 2.2.1. Then Singer and Wu computes the optimal orthogonal transformation between bases. It can be regarded as a numerical approximation to the parallel transport operator between the tangent spaces. \\\\
Firstly, we collect the base matrix $O_i \in \mathcal{V}_d(\mathbb{R}^N)$ by the local PCA. It is an approximation to the orthogonal basis for $T_{x_i}M$. Pick a small distance parameter $\epsilon$ but the parameter $\epsilon$ is much bigger than the scale parameter $\epsilon_{\text{PCA}}$ in the local PCA method (see Section 2.2.1). If points $x_i$ and $x_j$ satisfies $||x_i-x_j||_{\mathbb{R}^D}^2<\epsilon$, we compute the approximation matrix $O_{ij}$ as a transport vectors from $T_{x_j}M$ to $T_{x_i}$ by 
\begin{align}
      O_{ij}:=\arg \min_{O\in O(d)}||O-O_i^T\cdot O_j||_{HS},
\end{align}
where $O(d)$ denotes the orthogonal group and the norm in (2.3.6) is called the Hilbert-Schmidt norm, i.e.
\begin{align*}
||A||_{HS}:=tr(A\cdot A^T).
\end{align*}
The optimizer of (2.3.6) is given by$U\cdot V^T$, where $U$ and $V$ are the left and right eigenvectors of  $O_i^TO_j$ by SVD. \\\\
Now, we consider a weighted graph. The node of the graph is data and the edge is connected if two nearby data has $\epsilon$ Euclidean distance. The weights $w_{ij}$ of the graph is given by
\begin{align*}
       w_{ij}=K\left(\frac{||x_i-x_j||_{\mathbb{R}^N}}{\sqrt{\epsilon}}\right),
\end{align*}
where $K(\cdot)$ is the Gaussian kernel.\\\\
Construct an $N\times N$ block matrix $S$. Each block of matrix $S$ is $d\times d$ submatrix, i.e.
\begin{align*}
 S(i,j)= \begin{cases}
                  w_{ij}O_{ij} & w_{ij}>0\\
                  0_d             & w_{ij}=0,
            \end{cases}
\end{align*}
The diagonal $d\times d$ block matrix $D$ is defined by 
\begin{align*}
       D(i,i)=w(i)\cdot I_d.
\end{align*}
Note that the matrix $S$ and matrix $D$ has the same size $Nd\times Nd$. Consider a symmetric form $\bar{S}$ for matrix $S$ as 
\begin{align*}
            \bar{S}=D^{-\frac{1}{2}} S D^{-\frac{1}{2}}.
\end{align*}
Then we decompose $\bar{S}$ by the spectral decomposition with the eigenvalues $\lambda_1,\dots, \lambda_{Nd}$ in a descending order and the corresponding eigenvectors are $\phi_1,\dots, \phi_{Nd}$. We have 
\begin{align*}
         \bar{S}(i,j)&=\sum_{k=1}^{Nd}[\lambda_k \cdot \phi_k(i)]\phi_k(j)^T\\
         \bar{S}^{2t}(i,j)&=\sum_{k=1}^{Nd}[\lambda_k^{2t}\cdot \phi_k(i)]\phi_k(j)^T,
\end{align*}
where $\phi_k(i) \in \mathbb{R}^d$ for $i, j=1,\dots,N$ and $k=1,\dots,Nd$.\\\\
Then the vector diffusion map $[\mathcal{VD}_t]: \mathbb{R}^N\rightarrow \mathbb{R}^{(Nd)^2}$ is given by
\begin{align*}
     [\mathcal{VD}_t](x_i):=\left( (\lambda_k\cdot \lambda_m)^t \cdot \langle \phi_k(i), \phi_m(i)\rangle_{\mathbb{R}^d}\dots \right)_{k, m=1}^{Nd}.
\end{align*}
\textbf{Note:}
\begin{itemize}
\item The vector diffusion map $[\mathcal{VD}_t](x_i)$ is invariant to the choice of the basis of $T_{x_i} M$ since the inner product $\langle \phi_k(i), \phi_m(i)\rangle_{\mathbb{R}^d}$ preserves the value under the orthogonal transformations.
\item The inner product $\langle [\mathcal{VD}_t](x_i), [\mathcal{VD}_t](x_j)\rangle_{\mathbb{R}^{(Nd)^2}}$ gives a vector diffusion distance on $\mathbb{R}^{(Nd)^2}$ and we have
\begin{align*}
        \langle [\mathcal{VD}_t](x_i), [\mathcal{VD}_t](x_j)\rangle_{\mathbb{R}^{(Nd)^2}} = ||\bar{S}^{2t}(i,j)||_{HS}^2
\end{align*}
\end{itemize}
\clearpage
\section{Comparison}
The comparison of the data reduction algorithms includes several factors as follows \cite{wittman}:
\begin{itemize}
\item Cspeed: The computational speed for the algorithms to solve the problems. (Very fast $>$ Fast $>$ Median $>$ Slow $>$ Very slow) 
\item Geometry: Whether the new representation of the data holds the original geometric relations. (Yes/ No)
\item Noise: The ability for the algorithm to deal with the noise data which is meaningless data caused by hardware failures, programming errors or other reasons. (Good $>$ Median $>$ Bad)
\item Distribution: The performance of the algorithm under the different probability distribution. (Any type/ Uniform)
\item Clustering: The ability of the algorithm to hold the data in the same group after the reduction. (Good $>$ Median $>$ Bad)
\item Hdimension: The ability of the algorithm to deal with data of a high dimension. (Very good $>$ Good $>$ Median $>$ Bad $>$ Very Bad)
\item Sensitivity: The algorithm is sensitive to changing of the parameters. (Very $>$ Yes $>$ No) 
\end{itemize}
The factors listing above is suggested by most of manifold learning methods in numerical experiments. In the following section, we make two tables. Table \ref{Tab1} records the performance of seven algorithms according to the above seven factors. We run the experiments on a MATLAB graphical user interface (GUI) called MANI GUI created by Todd Wittman. The datasets includes Swiss Roll, Toroidal Helix, Corner Planes and Punctured Sphere. In particular, we choose three dimensional data sets and reduce them to the two dimensional plane. Table \ref{Tab2} records the decomposed matrices of each algorithm in the first column. The decomposed matrix means the matrix we need to decompose by SVD method in each algorithm. The second column records the truncated coordinates of the new representations of data in feature space. We take the $d$ coordinates except the Vector Diffusion Map algorithm. 

\begin{sidewaystable}
\renewcommand{\arraystretch}{1.5}
\centering
\begin{tabular}{ |p{4cm}||p{2cm}|p{2cm}|p{2cm}|p{2cm}|p{2cm}|p{2cm}|p{2cm}|p{2cm}|} 
 \hline
Algorithm Name & Cspeed & Geometry & Noise & Distribution & Clustering & Hdimension & Sensitivity\\
 \hline
PCA  &Very fast&No&Good&Any type&Good&Good&No \\
 \hline
ISOMAP &Very slow &Yes&Median&Any type&Good&Best&Yes \\   
\hline    
LLE         &Fast &Yes&Bad&Any type&Good&Very bad&Yes  \\
\hline
EigenMap &Fast &Yes&Good&Uniform&Bad&Bad&Yes\\
\hline
Hessian EigenMap &Slow&Yes&Good&Any type&Bad&Very bad&Yes \\
\hline
LTSA  &Fast &Yes&Good&Any type&Good&Good&Yes \\
\hline 
Diffusion Map &Fast &No &Good &Any type &Good& Good&Very \\
\hline 
\end{tabular}
\caption{\label{Tab1} Comparison of the Algorithms}
\end{sidewaystable}

\begin{table}
\renewcommand{\arraystretch}{1.5}
\centering
\begin{tabular}{ |p{4cm}||p{5cm}|p{7cm}|} 
 \hline
Algorithm Name & Decomposed Matrix &Truncated Coordinators\\
 \hline
PCA & $\Sigma_N$ & ${[v_1,\dots,v_d]}^T\times (x_i-\bar{x})$ \\
 \hline
ISOMAP & $G_g=C_N\cdot (-\frac{1}{2}d_G^2(i, j))\cdot C_N$ &  $(\sqrt{\lambda_1}\cdot v_1(i),\dots,\sqrt{\lambda_d}\cdot v_d(i))$ \\   
\hline    
LLE     &   $M=(I-W)^T\cdot (I-W)$   &  $\left(v_1(i),\dots,v_d(i)\right)$ \\
\hline
EigenMap & $L=D-W$ & $\left(f_1(i), \dots, f_d(i)\right)$  \\
\hline
Hessian EigenMap & $H:=\sum_i H\left(N(i),N(i)\right)$ & $\left(v_1(i),\dots,v_d(i)\right)$\\
\hline
MVU   &  $K^*$ &$\left(\sqrt{\lambda_1}\cdot v_1(i),\dots,\sqrt{\lambda_d}\cdot v_d(i)\right)$ \\
\hline
LTSA  & $S\cdot WW^T\cdot S^T$ &$\left(\sqrt{\lambda_1}\cdot v_1(i),\dots,\sqrt{\lambda_d}\cdot v_d(i)\right)$ \\
\hline 
Diffusion Map &  $P^t$  & $\left(\lambda_1^t\cdot \phi_1(u), \dots, \lambda_n^t\cdot \phi_d(u)\right)$ \\
\hline 
Vector Diffusion Map & $\bar{S}=D^{-\frac{1}{2}} \cdot S \cdot D^{-\frac{1}{2}}$ & $\left( (\lambda_k\lambda_m)^t \cdot \langle \phi_k(i), \phi_m(i)\rangle_{\mathbb{R}^d} \dots \right)_{k, m=1}^{Nd}$ \\
\hline 
\end{tabular}
\caption{\label{Tab2}Manifold Learning Algorithm}
\end{table}

\chapter{Convergence Issues}
In differential geometry, a topological manifold is an abstract set (topological space) which is homeomorphic to the Euclidean space locally. People assign some differential structure to it and then it becomes a differential manifold. In 1851, Riemann firstly described his idea of Riemannian metric in his defense for the habilitation, which began a new era in geometry. However, it is not until Whitney's work in 1936 that mathematicians got a clear understanding of the abstract manifold, which is just a submanifold embedded in Euclidean space \cite{whitney1934analytic}. Later in 1954 and 1956, Nash discovered the amazing result of the $C^1$ and $C^k$ isometric embedding of Riemannian manifolds into some Euclidean space (\cite{nash1956imbedding}, \cite{nash1954c1} respectively).\\\\
In the EigenMap, Diffusion Map and Vector Diffusion Map, we assume the embedding is (Riemannian) isometric. i.e. suppose $\iota=(\iota_1,\dots,\iota_D) : M^d \hookrightarrow \mathbb{R}^D$ is a smooth Riemannian submanifold isometrically embedded in $\mathbb{R}^D$ with the induced metric $g_M$ from the canonical metric on ($\mathbb{R}^D$, can). Then the isometric embedding map $\iota$ satisfies
\begin{align*}
        g_{ij}=\langle D\iota(\frac{\partial}{\partial x_i}), D\iota(\frac{\partial}{\partial x_j})\rangle_{\mathbb{R}^D}.
\end{align*}
In local coordinates, this is
\begin{align}
     g_{ij}= \sum_{r=1}^D \frac{\partial \iota_r}{\partial x_i} \cdot \frac{\partial \iota_r}{\partial x_j}.
\end{align}
In differential geometry, the isometric embedding problem is to find a one-to-one function $\iota$ satisfying equation (3.0.1). Our data is isometrically embedded into Euclidean space and thus the aim is to find a representation of the data in the feature space which is low dimensional Euclidean space. In general, we collect a finite number of samples of the high-dimensional data and assume they are on or near a manifold with an unknown dimension. Then, we reduce the data to a low dimension as written in Chapter 2. In this section, we talk about the convergence issue of three manifold learning algorithms. Notice that we always use the Einstein summation convention in this chapter.\\\\
In general, the strategy of the convergence issue has two steps. The first step is to construct a convergence relationship between the averaging operator and the (normalized/unnormalized) discrete graph Laplacian, since the information we can collect is from the discrete graph structure generated by the finite data samples in the sample space. The second step is to show the averaging operator converges to the Laplace-Beltrami operator as the time parameter goes to $0^+$. This step mainly utilizes the basic technique of the heat kernel estimation on manifolds and general operator theory. With these two steps, we can reduce the bound of the difference between the discrete graph Laplacian and the Laplace-Beltrami operator which is called the \emph{bias} term and \emph{variance} term. We will discuss this further in Section 3.3.
\section{Embedding Relation}
This section is about the pure geometric results which expose the geometric relations between the intrinsic quantities on a manifold and the extrinsic Euclidean quantities in the ambient space. These relations are very important technical results often used in replacing the Euclidean distance by the geodesic distance on a manifold which includes some higher-order terms like curvature term. It is also helpful in deducing an asymptotic expansion for the (heat) operator.  \\\\
Let $\iota: M^d \hookrightarrow \mathbb{R}^D$ be a connected compact smooth Riemannian submanifolds isometrically embedded in $\mathbb{R}^D$ with the induced metric $g$ from the canonical metric on ($\mathbb{R}^D$, can). Recall that in the Riemannian normal coordinate system, we can expand the Taylor series of the metric to higher order terms at any point $p \in M$, i.e.
\begin{align}
  g_{ij}(x(p))\sim \delta_{ij}-\frac{1}{3}R_{iklj}x^kx^l-\frac{1}{6}R_{iklj,m} x^kx^lx^m,
\end{align}
where the curvature tensor $R$ denoted as 
\begin{align*}
R_{iklj}&:=g\big(R(\frac{\partial}{\partial x_l},\frac{\partial}{\partial x_j})\frac{\partial}{\partial x_k}, \frac{\partial}{\partial x_i}\big),\\
R_{iklj,m}&:=\nabla_{\frac{\partial}{\partial x_m}} R_{iklj}.
\end{align*}
Let $(W_p, \text{exp}_p)$ be the normal coordinate system around $p\in M^d$, i.e.
\begin{align*}
\text{exp}_p : W_p &\rightarrow M\\
                      w & \mapsto \gamma_w(1),
\end{align*}
where $\gamma_w$ is the geodesic of $M$ with $c(0)=p$ and $\dot{c}(0)=w$, and $W_p:=\{w\in T_p M: \gamma_w $ is defined on $[0,1]\}$.\\\\
One relationship based on (3.1.1) between the geodesic distance on a manifold and the Euclidean distance in the embedding space is described in \cite[Prop 1]{smolyanov2000brownian}. Note that $d\text{exp}_p$ is nonsingular around $p$. Then there exists a neighborhood $U_p \subset M$ such that $\text{exp}_p^{-1}$ is diffeomorphical onto a neighborhood $V \subset T_p M\cong \mathbb{R}^d$. Thus, we have the following embedding relation, for all $v \in V$
\begin{align*}
        ||v||_{\mathbb{R}^d}^2 \overset{(1)}{=} d_M^2(p,\text{ exp}_p(v))\overset{(2)}{=} ||\iota \circ \text{exp}_p(v)-\iota(p)||_{\mathbb{R}^D}^2+ O(||v||_{\mathbb{R}^d}^4),
\end{align*}
where equality (1) is deduced from the definition of exponential map, and equality (2) follows from the properties of the normal coordinate system.\\\\
For any $v, w \in V \subset T_p M$, we have more specific embedding equalities mentioned in \cite[Lemma B.7 B.8 \& B.9]{singer2012vector}
\begin{align}
      \iota\circ \text{exp}_p(v) - \iota(p) &= d\iota(v)+\frac{1}{2}\Pi(v, v) +\frac{1}{6}\Pi(v,v) + O(||v||_{\mathbb{R}^d}^4), \\
   d [\iota\circ \text{exp}_p](v)(w)-d[ \iota\circ \text{exp}_p ](0)(w) &= \Pi(v, w)+\frac{1}{6}\nabla_v \Pi(v,w)+\frac{1}{3}\nabla_w \Pi(v,v) + O(||v||_{\mathbb{R}^d}^3),
\end{align}
where $\Pi$ is the second fundamental form of the embedding manifolds $M^d\hookrightarrow \mathbb{R}^D$.
Equalities (3.1.2) and (3.1.3) provide a new estimate of $||v||_{\mathbb{R}^d}$: suppose $h:=||\iota\circ \text{exp}_p(v) - \iota(p)||_{\mathbb{R}^D}$ and $\theta:=v/||v||$. For a small $||v||_{\mathbb{R}^d}^2$, we have 
\begin{align*}
        ||v||_{\mathbb{R}^d} =||\iota \circ \text{exp}_p(v)-\iota(p)||_{\mathbb{R}^D}+\frac{1}{24}||\Pi(\theta,\theta)||_{T_p\mathbb{R}^D}^2\cdot h^3 + O(h^4).
\end{align*}
In the normal coordinate system $(U_p, \text{ exp}_p; \hspace*{0.5em} x_1,\dots, x_d)$, let $\{\frac{\partial}{\partial x_l}\}_{l=1}^d$ be the normal coordinate vector field on $U_p$. For any $q \in U_p$ with $q:=\text{exp}_p(v)$, where $v \in V \subset T_p M$, we have a relation of orthogonal bases under the parallel transportation from $p$ to $q$
\begin{align*}
\iota_* P_{q,p} \frac{\partial}{\partial x_l}(p) =& \iota_*  \frac{\partial}{\partial x_l}(p) + ||v||\cdot \Pi(\theta,  \frac{\partial}{\partial x_l}(p)) +  \frac{||v||^2}{6} \cdot \nabla_{\theta} \Pi(\theta,  \frac{\partial}{\partial x_l}(p)) \\
&+\frac{||v||^2}{3} \cdot \nabla_{ \frac{\partial}{\partial x_l}(p)} \Pi(\theta,\theta) - \frac{||v||^2}{6} \cdot \iota_*P_{q,p}\big(R(\theta, \frac{\partial}{\partial x_l}(p) )\theta\big)+ O(||v||_{\mathbb{R}^d}^3).
\end{align*} 
\section{Formulation of the Algorithms via Operator Theory}
In Chapter 2, we has derived the three algorithms EigenMap, Diffusion Map and Vector Diffusion Map in the language of matrix theory so that people can compute eigenvalues and eigenfunctions of the discrete graph Laplacian generated by the data samples. In this chapter, we introduce a theoretical formulation for these three algorithms in the language of operator theory on manifolds. We will not mention the conventions and notations in Chapter 2 agian. The references include EigenMap \cite{belkin2003problems, belkin2003laplacian, von2008consistency}, Diffusion Map \cite{lafon2004diffusion, coifman2006diffusion} and Vector Diffusion Map \cite{singer2012vector}. 
\subsection{EigenMap (2003, M. Belkin and P. Niyogi)}
Suppose $f\in C^2(M)$. We have an inequality that bounds how far $f$ can map two points $x, y \in M$ from each other. This is 
\begin{align*}
   |f(y)-f(x)| \leq ||\nabla f(x)||_{T_x M}\cdot ||y-x||_{\mathbb{R}^D}+o\big(||y-x||_{\mathbb{R}^D} \big)
\end{align*}  
The aim of the EigenMap is to look for a map satisfying 
\begin{align}
     \mathop{\argmin}_{||f||_{L^2(M)}=1} \int_M ||\nabla f(x)||_{T_p M}^2,
\end{align}
which means that the close data will be as close as possible after the mapping. \\\\
Note that 
\begin{align*}
     & \int_M ||\nabla f(x)||_{T_p M}^2\\
   =& \int_M \langle \nabla f, \nabla f \rangle_{T_p M}^2 \\
   =& \int_M \mathcal{L}(f)\cdot f\\
   =& \lambda \cdot \int_M f^2 \\
   =& \lambda \cdot ||f||_{L^2(M)}.
\end{align*}
Thus, to minimize (3.2.1) is to spectral decompose the Laplace-Beltrami operator $\mathcal{L}$, which only picks the discrete eigenvalues $0=\lambda_0 \leq \lambda_1 \leq \lambda_2 \dots \rightarrow \infty$. \\\\
Note that there is a big relationship between the heat equation and the Laplace-Beltrami operator on manifolds. Thus, we introduce the heat equation here with the aim of approximating the heat operator by the graph Laplacian. Consider the homogeneous heat equation on the manifolds 
\begin{align*}
   (\Delta+\partial_t )u(x,t)&=0 \hspace*{2em} (x,t)\in M\times(0,\infty),\\
   u(x,0)&=f(x) \hspace*{1.2em} \hspace*{1em}  x\in M.
\end{align*}
Then the general solution is given by 
\begin{align*}
   u(x, t)= \int_{M(y)} u_y(x, t)f(y),
\end{align*}
where $u_y(x, t)$ is the fundamental solution. \\\\
Note that 
\begin{align*}
 [\mathcal{L}f](x):&=(\Delta+\partial_t )f(x)\\
                         &= (\Delta+\partial_t )u(x,0)\\
                         &=- [\frac{\partial}{\partial t} \int_{M(y)} u_y(x, t)f(y)]\big|_{t=0}\\
                         &\overset{(*)}\approx \lim_{t\rightarrow 0^+}-[(4\pi t)^{-\frac{d}{2}}\int_{M(y)} e^{-\frac{d_g^2(x,y)}{4t}}f(y)-f(x)]/t.
\end{align*}
The approximation step $(*)$ follows from two facts. One is that the fundamental solution $u_y(x,t)$ tends to the Dirac distribution $\delta_y(x)$ as $t$ tends to $0^+$, i.e.
\begin{align*}
                   \lim_{t\rightarrow{0^+}} \int_{M(y)} u_y(x,t)f(y) = f(x) \hspace*{0.5em} \text{ for all } x \in M.
\end{align*}
The other follow from the fact that the fundamental solution $u_y(x,t)$ has an expansion when data $x$ and $y$ are close on the manifold and time $t$ is very small, i.e.
\begin{align*}
        u_y(x,t)\approx (4\pi t)^{-\frac{d}{2}}e^{-\frac{d_g^2(x,y)}{4t}},
\end{align*}
which is known as Varadhan's large deviation formula relating the heat kernel and geodesic distance on a Riemannian manifold \cite{saloff2010heat}.\\\\ 
In practice, since the number of the sample is finite, we can only construct the discrete graph Laplacian $\bar{\mathcal{L}}$ based on the data information, i.e.
\begin{align*}
[\bar{\mathcal{L}}f](x_i) \approx \lim_{t\rightarrow 0^+}-[\frac{1}{N}(4\pi t)^{-\frac{d}{2}}\cdot \sum_{x_j} e^{-\frac{d_g^2(x_i-x_j)}{4t}}f(x_j)-f(x_i)]/t
\end{align*}
where $x_j \in \{x: 0<d_g(x_i-x)<\epsilon\}$.
\subsection{Diffusion Maps (2004, S. Lafon)}
In this section, the formulation of the algorithm is still in a discrete sense but we define the Diffusion Map using the language of operator theory on manifolds. Suppose the data samples are distributed on the measure space $(M, \mathcal{B}(M), \omega_g)$ of the manifold $(M, g)$ and we use the kernel $K(x,y)$ to represent the dissimilarity between two data point $x$ and $y$. In general, we need the kernel function to be symmetric, positivity-preserving and positive semi-definite, i.e.
\begin{itemize}
\item Symmetric: $K(x,y)=K(y,x)$
\item Positivity-preserving: $K(x,y)\geq 0$ for any $x$ and $y$ in $M$
\item Positive semi-definite: for all bounded function $f$ defined on $M$
         \begin{align*}
               \int_{M(x)} \int_{M(y)} K(x,y) f(x) f(y) \geq 0.
         \end{align*}
\end{itemize}
The kernel function satisfying the conditions above is called the \emph{admissible kernel}. Then a routine in the kernel-based method is to normalize the kernel by introducing a constant $v^2(x)$ defined as 
\begin{align*}
   v^2(x) :=\int_{M(y)} K(x,y).
\end{align*}
The normalizing routine of the admissible kernel is to transform the original kernel to a Markov kernel by $\tilde{a}(x,y):=\frac{K(x,y)}{v^2(x)}$ on the data graph. Since the Markov chain exists for any Markov kernel \cite[Prop 1.5]{grigoryan2009analysis}, there always exists a random walk on $M$ for any admissible kernel. Thus we have a computational version of this algorithm starting directly from the random walk in Chapter 1. 
\\\\
Define the discrete averaging operator $\mathcal{A}: L^2(M)\rightarrow L^2(M)$ 
\begin{align*}
 [\mathcal{A}f](x)=\int_{M(y)} a(x,y)f(y) 
\end{align*}
where the entries of the operator $\mathcal{A}$ are given by
\begin{align*}
a(x,y):=\frac{K(x,y)}{v(x)v(y)}.
\end{align*}
According to the construction above, the averaging operator $\mathcal{A}$ is bounded, symmetric and positive semi-definite with the supremum norm 1 on $L^2(M, \omega_g)$. In general, we say a densely defined operator $\mathcal{A}:D(\mathcal{A})\subset \mathcal{H}\rightarrow \mathcal{H}$ is \emph{symmetric} if $\mathcal{A}^*$ is an extension of $\mathcal{A}$, i.e. $\mathcal{A}\subset \mathcal{A}^*$ and \emph{self-adjoint} if $\mathcal{A}=\mathcal{A}^*$. If $\mathcal{A}$ is continuous and has domain $D(\mathcal{A})=\mathcal{H}$, symmetry of $\mathcal{A}$ implies self-adjointness of $\mathcal{A}$. Since the averaging operator $\mathcal{A}$ is bounded and self-adjoint, the spectral decomposition theorem implies
\begin{align*}
a(x,y)=\sum_{i\geq 0}^{N-1} \lambda_i \phi_i(x) \phi_i(y),
\end{align*} 
where the eigenvalue $\lambda_i$ are non-increasing and non-negative satisfying
\begin{align*}
\mathcal{A}\phi_i(x)=\lambda_i \phi_i(x).
\end{align*}
Moreover, the $t$ th-step kernel $\mathcal{A}^t$ satisfies
\begin{align*}
a^{t}(x,y)=\sum_{i\geq 0}^{N-1} \lambda_i^m \phi_i(x) \phi_i(y).
\end{align*}
\\\\
Finally, the diffusion map $[\mathcal{D}]: M \rightarrow l^2(\mathbb{N})$ is given by 
\begin{align*}
       [\mathcal{D}]: x\mapsto \left(\phi_0(x), \dots, \phi_{N-1}(x)\right),
\end{align*}
and a family of semi-metric $\{ d_{\text{Diffusion, t}} \}_{t\geq 1}$ on $M$ is defined by
\begin{align*}
d_{\text{Diffusion, t}} (x, y):=\sqrt{a^{t}(x,x)-2\cdot a^{t}(x,y)+a^{t}(y,y)}.
\end{align*} 
\subsubsection{Note}
\begin{itemize}
\item If the kernel function $K(x,y)$ is strictly positive definite, the $\{d_{\text{Diffusion, t}} \}_{t\geq 1}$ is the true metric. 
\item $d_{\text{Diffusion, 2t}}^2(x, y)$ is the Euclidean distance between the columns of indices $x$ and $y$ of $\mathcal{A}^m$, i.e.
\begin{align*}
d_{\text{Diffusion, 2t}}^2(x, y) &= \int_\Gamma |a^{t}(x,z)-a^{t}(y,z)|^2 d\mu(z) = ||a^{t}(x,\cdot)-a^{t}(y,\cdot)||^2.
\end{align*}
\item The diffusion metric $d_{\text{Diffusion, t}}$ is equal to the weighted Euclidean distance in the embedding space with the weights $\lambda_i^t$, i.e.
\begin{align*}
d_{\text{Diffusion, t}}^2 (x, y)&= \sum_{i\geq 0}^{N-1} \lambda_i^t \big(\phi_i(x)-\phi_i(y)\big)^2.
\end{align*}
\end{itemize}
\subsection{Vector Diffusion Maps (2011, A.Singer and H.-T. Wu)}
The Laplacian operator $\Delta$ can be extended to act on the tangent bundle $TM$ of the Riemannian manifold $(M^d, g)$. It is defined as the trace of the second covariant derivative with the metric $g$. Mathematically, for any tensor field $F \in \Gamma_l^k(M)$
\begin{align*}
\Delta F:= \text{tr } \nabla^2 F
\end{align*} 
where $\nabla^2$ is the second covariant derivative, i.e. for any vector field $X$ and $Y$
\begin{align*}
\nonumber
\nabla^2 F(\omega^1,\dots, \omega^l, Y_1, \dots, Y_k, X, Y)=& \nabla_X \big( \nabla_Y F(\omega^1,\dots, \omega^l, Y_1, \dots, Y_k)\big)-\\
 & \nabla_{\nabla_X Y}F(\omega^1,\dots, \omega^l, Y_1, \dots, Y_k).
\end{align*}
Thus we have
\begin{align*}
       \Delta F = \sum_{i=1}^d \nabla^2 F(\dots, E_i, E_i).
\end{align*}
\\\\
In the classical elliptic theory, $e^{t\Delta}$ has kernel of the following form 
\begin{align*}
      k_t(x,y)=\sum_{i=0}^\infty e^{-\lambda_it} X_i(x) \cdot \overline{X_i(y)}
\end{align*}
where $0\leq \lambda_0\leq \lambda_1 \dots $ and $X_i$ are the associated eigenvector fields satisfying 
\begin{align*}
     \Delta X_i = -\lambda_i X_i.
\end{align*}
Moreover, the eigenvector fields $X_i$ for the Laplacian operator $\Delta$ form an orthonormal basis of $L^2(TM)$. Note that the Laplacian defined on a tangent bundle holds the positive trace and thus all the signs for the equations are opposite to those in Appendix C \& D. \\\\
Define the vector diffusion map $[\mathcal{VD}_t]: M \rightarrow l^2$
\begin{align*}
      [\mathcal{VD}_t]: x \mapsto \big(e^{\frac{(\lambda_i+\lambda_j)t}{2}}\langle X_i(x), X_j(x)\rangle \big)_{i,j=0}^\infty
\end{align*}
\subsubsection{Note}
\begin{itemize}
\item The vector diffusion map is a diffeomorphic embedding of $M$ into $l^2$. 
\item The vector diffusion distance is given by
\begin{align*}
d_\text{VDM,t}(x,y):=\big|\big| [\mathcal{VD}_t](x)- [\mathcal{VD}_t](y)\big|\big|_{l^2},
\end{align*}
which has an asymptotic expansion. For any $x, y \in M$ with $y=\text{exp}_x v$, where $v\in T_x M$ and $||v||^2 < t< 1$, then 
\begin{align*}
              d_\text{VDM,t}^2(p,q) =\frac{d}{4\pi^{d}} \frac{||v||^2}{t^{d+1}}+ O\left(\frac{1}{t^{d}}\right).
\end{align*}
\end{itemize}
\section{Approximation to the Laplace-Beltrami Operator}
In this section, we discuss the core techniques of the convergence issue for the three algorithms. In the literature, the difference between the graph Laplacian operator and the Laplace-Beltrami operator contains two parts, the \emph{variance term} and the \emph{bias term}. Variance term establishes the convergence of the graph Laplacian to some continuous operator called the averaging operator and the bias term establishes the convergence of this continuous operator to the Laplace-Beltrami operator on manifold. The two terms are related to the time parameter and the quantities of the data samples. The following diagram exposes this relation
\begin{align*}
     \text{Graph Laplacian } \bar{\mathcal{L}}_{\epsilon, N} \overset{N\rightarrow \infty}{\longrightarrow} \text{Averaging } \mathcal{A}_\epsilon \overset{\epsilon \rightarrow 0^+}{\longrightarrow}\text{Laplace-Beltrami } \mathcal{L} (\text{or }\Delta). 
\end{align*}  
Thus, to show the convergence is to reduce the \emph{variance} term and the \emph{bias} term. In the following section, we will show the idea of the convergence issue of each algorithm one by one. The reference includes EigenMap \cite{belkin2003problems, belkin2003laplacian, von2008consistency}, Diffusion Map \cite{lafon2004diffusion, coifman2006diffusion} and Vector Diffusion Map \cite{singer2012vector}. 
\subsection{EigenMap (2003, M.Belkin and P.Niyogi)}
Let the Laplace-Beltrami operator $\mathcal{L}^t$ at time $t$ be
\begin{align*}
  [\mathcal{L}^t f](p) = (4\pi t)^{-\frac{d+2}{2}}\cdot \int_{M(q)} e^{-\frac{||p-q||^2}{4t}}[f(p)-f(q)]
\end{align*}
and, the graph Laplacian operator $\mathcal{L}_N^t$ at time $t$
\begin{align*}
  [\mathcal{L}_N^t f](p) = \frac{(4\pi t)^{-\frac{d+2}{2}}}{N}\cdot \sum_{x_i} e^{-\frac{||p-x_i||^2}{4t}}[f(p)-f(x_i)]
\end{align*}
where $x_i\in \{x:0<||p-x||<\epsilon\}$.\\\\
The main structural result of this algorithm is as follows
\begin{mdframed}[backgroundcolor=blue!5] 
\vskip 0.2cm
\textbf{Main Result:} Suppose the data $\{x_i\}$ is i.i.d and uniformly distributed on $M$. Then
\begin{align}
  \text{Eigen } \mathcal{L}_N^t \overset{N\rightarrow \infty}{\longrightarrow} \text{Eigen } \mathcal{L}^t \overset{t\rightarrow 0^+}{\longrightarrow}\text{Eigen } \mathcal{L}
\end{align}
where the first approximation means that the eigenfunctions and the associated eigenvectors of $\mathcal{L}_N^t$ approach to the ones of $\mathcal{L}^t$ almost surely as the number of samples blows up, and, the second means that the eigenvalues and the associated eigenfunctions of $\mathcal{L}^t$  approach to those of  $\mathcal{L}$ as time $t$ approaches to $0^+$.
\end{mdframed}
\vskip 0.5cm
Let $\{\lambda_i\}$, $\{\lambda_{i}^t\}$ and $\{\lambda_{N,i}^t\}$ be the eigenvalues for $\mathcal{L}$, $ \mathcal{L}^t  $ and $\mathcal{L}_N^t$ respectively. And, $\{\phi_i\}$, $\{\phi_{i}^t\}$ and $\{\phi_{N,i}^t\}$ are the associated eigenvectors with respect to $\{\lambda_i\}$, $\{\lambda_{i}^t\}$ and $\{\lambda_{N,i}^t\}$ respectively. Then the main result above says for all $i$, we have almost surely
\begin{align*}
 \lim_{t\rightarrow 0} \lim_{N\rightarrow \infty} &|\lambda_{N,i}^t-\lambda_i|=0,\\
  \lim_{t\rightarrow 0} \lim_{N\rightarrow \infty} &||\phi_{N,i}^t-\phi_i||_{L^2(M)}=0. 
\end{align*}
\emph{Basic Idea of Proof:}\\\\
Denote the heat operator $\mathcal{H}^t$ at time $t$ as
\begin{align*}
      [\mathcal{H}^tf](x):=\int_{M(y)} u_y(x,t) f(y),
\end{align*}
and recall that the Laplace-Beltrami operator $\mathcal{L}$ is
\begin{align*}
[\mathcal{L}f](x)&= \lim_{t\rightarrow 0^+} [f(x)-\int_{M(y)} u_y(x, t)f(y)]/t \\
                        &=\lim_{t\rightarrow 0^+} [\frac{1-\mathcal{H}^t}{t}f](x).
\end{align*}
The approximate operator $\frac{1-\mathcal{H}^t}{t}$ does not converge uniformly to $\mathcal{L}$ in the sense that\begin{align*}
 \sup_{||f||_{L^2}=1} ||[\frac{1-\mathcal{H}^t}{t}f]-[\mathcal{L}f]||_{L^2(M)} \not\rightarrow 0.
\end{align*}
Consider a perturbation operator $\mathcal{R}^t: L^2(M)\rightarrow L^2(M)$ at time $t$
\begin{align*}
\mathcal{R}^t:=\frac{1-\mathcal{H}^t}{t}-\mathcal{L}^t.
\end{align*}
By the result in \cite[Theorem 4.1]{belkin2003laplacian}, we have
\begin{align*}
           \lim_{t\rightarrow 0^+} \sup_{||f||_{L^2}=1} \frac{\langle \mathcal{R}^tf, f \rangle}{\langle \frac{1-\mathcal{H}^t}{t}f, f  \rangle}=0,
\end{align*}
which implies the eigenvalues and associated eigenfunctions of the approximate operator $\frac{1-\mathcal{H}^t}{t}$ converge to the ones of $\mathcal{L}^t$. Observe that $\mathcal{L}$ and $\frac{1-\mathcal{H}^t}{t}$ share the same eigenfunctions. Thus we get the second approximation of the main result, i.e. Eigen $\mathcal{L}^t \overset{t\rightarrow 0^+}{\longrightarrow}\text{Eigen } \mathcal{L}$. The variance term part can be found in \cite{von2008consistency}.
\subsection{Diffusion Maps (2004, S. Lafon)}
Consider the rotation invariant kernel, i.e.
\begin{align*}
k(x, y)=h(||(x,y)||^2)
\end{align*}
where the map $u\mapsto h(u^2)$ must be chosen as the Fourier transform of a finite positive measure by the Bochner's theorem which guarantees the positivity of the kernel \cite{bochner1941hilbert}. Define the $\epsilon$-kernel $k_\epsilon(x,y)$ by
\begin{align}
       k_\epsilon(x, y):= h\left(\frac{||x-y||^2}{\epsilon}\right).
\end{align}
\\\\
Suppose we are given a class $\{E_K\}_{K>0}$ of functions $f \in C^\infty(M)$ satisfying 
\begin{itemize}
\item For all multiple index $\alpha=(\alpha_1,\dots, \alpha_d)$
        \begin{align*}
                ||\frac{\partial^{|\alpha|} f}{\partial x_1^{\alpha_1} \dots \partial x_d^{\alpha_d}}||_{L^2(M)}\leq K^{|\alpha|} \cdot ||f||_{L^2(M)}.
        \end{align*}
\item For all $x \in \partial M$
        \begin{align*}
                     \frac{\partial f}{\partial r}(x) =0,
        \end{align*}
        where $r$ is a tangent vector at $x$ that is normal to $\partial M$.
\end{itemize}
One property of $\{E_K\}_{K>0}$ is as follows
\begin{align*}
        \overline{\bigcup_{K>0} E_K}=L^2(M).
\end{align*}
\\
In the original paper \cite{lafon2004diffusion}, the authors approximate the Laplace-Beltrami operator step by step. The first infinitesimal generator $G_\epsilon$ consists of curvature potential term. Then, the authors normalize the graph Laplacian, which works well for the uniformly distributed data on $M$. Eventually, they modify the kernel and the averaging operator by separating the geometry of $M$ from the distribution of the points and thus get it.
\subsubsection{Step 1: The infinitesimal generator $G_\epsilon$}
Consider an infinitesimal generator $G_\epsilon$
\begin{align*}
[G_\epsilon f](x):=\frac{1}{\epsilon^{\frac{d}{2}}} \int_{M(y)} k_\epsilon(x,y) f(y)
\end{align*}
which has an asymptotic expansion, i.e. for any $x\in M/\partial M$
\begin{align*}
[G_\epsilon f] (x) = \left( \int_{\mathbb{R}^d} h(||u||^2)du\right) f(x) + \frac{\epsilon \cdot \int_{\mathbb{R}^d}u_i^2 \cdot h(||u||^2)du}{2} \big(E(x)f(x)-\nabla f(x)\big) + O(\epsilon^{\frac{3}{2}})
\end{align*} 
where $a_i(x)$ is the curvature function of the coordinate geodesics at any point $x \in M$ and 
\begin{align*}
E(x)= \sum_{i=1}^d [a_i(x)^2-\sum_{j \neq i}a_i(x)\cdot a_j(x)].
\end{align*}
\subsubsection{Note}
\begin{itemize}
\item The infinitesimal generator $G_\epsilon$ combines the intrinsic geometry (the Laplace-Beltrami operator) and the extrinsic geometry (the curvature potential).
\item The curvature potential term $E(x)$ of the infinitesimal generator $G_\epsilon$ is zero when the manifold is a vector subspace of $\mathbb{R}^D$.
\end{itemize}
\subsubsection{Step 2: The Averaging Operator $\mathcal{A_\epsilon}$}
Suppose $p(y)$ is the density function for the measure $\omega_g$ on $M$, i.e. $\omega_g(y) =p(y) dy$. We then introduce the averaging operator $\mathcal{A_\epsilon}$ with the parameter $\epsilon$ via the general  normalizing routine of the graph Laplacian 
\begin{align*}
      [A_\epsilon f](x) := \frac{1}{v_\epsilon^2(x)}\int_{M(y)} k_\epsilon(x,y)f(y) \cdot p(y)dy,
\end{align*}
where 
\begin{align*}
v_\epsilon^2(x) = \int_{M(y)} k_\epsilon (x,y)\cdot p(y)dy,
\end{align*}
which has an asymptotic expansion: for any $f\in E_K$ and $x\in M/ \partial M$ 
\begin{align*}
[\mathcal{A}_\epsilon f](x) =f(x) + \frac{\epsilon \cdot  \int_{\mathbb{R}^d}u_i^2 \cdot h(||u||^2)}{2\cdot \int_{\mathbb{R}^d} h(||u||^2) } \cdot \left(\frac{\Delta p(x)}{p(x)}f(x)-\frac{\Delta (p\cdot f)(x)}{p(x)}\right) +O(\epsilon^{\frac{3}{2}}).
\end{align*}
\\\\
Define the graph Laplacian operator $\bar{\mathcal{L}}_\epsilon$ with the parameter $\epsilon$ as 
\begin{align*}
    \bar{\mathcal{L}}_\epsilon:=\frac{I-A_\epsilon}{\epsilon}.
\end{align*}
On the space $E_K$, we have 
\begin{align*}
\lim_{\epsilon \rightarrow 0^+} \bar{\mathcal{L}}_\epsilon = \mathcal{H},
\end{align*}
where 
\begin{align*}
[\mathcal{H} f](x) &:=\frac{ \int_{\mathbb{R}^d}u_i^2 \cdot h(||u||^2)}{2\cdot \int_{\mathbb{R}^d} h(||u||^2)} \cdot \left(\frac{\Delta (p\cdot f)(x)}{p(x)}-\frac{\Delta p(x)}{p(x)}f(x)\right)\\
                           &=\frac{ \int_{\mathbb{R}^d}u_i^2 \cdot h(||u||^2)}{2\cdot \int_{\mathbb{R}^d} h(||u||^2)} \cdot \left( \Delta f(x) +2 \langle \frac{\nabla p(x)}{p(x)}, \nabla f(x) \rangle \right).
\end{align*}
\subsubsection{Note}
\begin{itemize}
\item When the data is uniformly distributed over $M$, the limit operator $\mathcal{H}$ is a multiple of the Laplace-Beltrami operator on $M$.
\item The (weighted) graph Laplacian will not approximate the Laplace-Beltrami operator in the case of non-uniform densities. 
\end{itemize}
\subsubsection{Step 3: Modified Averaging Operator $\mathcal{A}_\epsilon$}
Construct the approximation density $p_\epsilon (x)$ with parameter $\epsilon$ by 
\begin{align*}
     p_\epsilon (x) = \int_{M(y)} k_\epsilon (x, y). 
\end{align*}
Now replace $k_\epsilon$ by
\begin{align*}
\tilde{k}_\epsilon(x,y):=\frac{k_\epsilon(x,y)}{p_\epsilon(x)\cdot p_\epsilon(y)},
\end{align*}
and let
\begin{align*}
v_\epsilon^2(x) = \int_{M(y)} \widetilde{k_\epsilon}(x,y). 
\end{align*}
Then the modified averaging operator is $\mathcal{A}_\epsilon : L^2(M)\rightarrow L^2(M)$,
\begin{align*}
[\mathcal{A_\epsilon}f](x):=\frac{1}{v_\epsilon^2(x)} \int_{M(y)} \tilde{k}_\epsilon(x,y) f(y)
\end{align*}
By the Laplace operator $\mathcal{L}_\epsilon$ with the parameter $\epsilon$ on $M$ is defined 
\begin{align*}
         \mathcal{L}_\epsilon := \frac{I-A_\epsilon}{\epsilon},
\end{align*}
which has an asymptotic expansion: for any $f\in E_M$ and $x\in M/\partial M$, then
\begin{align*}
     [\mathcal{A_\epsilon}f](x)=f(x)- \frac{\epsilon \cdot  \int_{\mathbb{R}^d}u_i^2 \cdot h(||u||^2)}{2\cdot \int_{\mathbb{R}^d} h(||u||^2)} \Delta f(x) +O(\epsilon^{\frac{3}{2}}),
\end{align*}
and on $E_K$, we have
\begin{align*}
    \lim_{\epsilon \rightarrow 0^+} \mathcal{L}_\epsilon = \frac{2 \cdot  \int_{\mathbb{R}^d}u_i^2 \cdot h(||u||^2)}{\int_{\mathbb{R}^d} h(||u||^2)} \cdot \Delta (:=\Delta_0).
\end{align*}
Moreover, since the modified operator $A_\epsilon$ is compact, we have
\begin{align*}
\lim_{\epsilon \rightarrow 0^+}  \mathcal{A_\epsilon}^{-\frac{t}{\epsilon}} = e^{-t\Delta_0}=\sum_{i\geq 0} \lambda_{\epsilon, i}^\frac{t}{\epsilon} P_{\epsilon, i},
\end{align*}
where $P_{\epsilon, i}$ is the orthogonal projector on the eigenspace associated to the eigenvalues $\lambda_{\epsilon, i}$. \\\\
In other words, the heat kernel $u_y(x,t)$ on $M$ can be approximated by $a_\epsilon^\frac{t}{\epsilon}(x,y)$ and we get 
\begin{align*}
      \lim_{\epsilon \rightarrow 0+} \lambda_{\epsilon, i}^\frac{t}{\epsilon} &= e^{-t\lambda_i},\\
      \lim_{\epsilon \rightarrow 0+} P_{\epsilon, i} &=P_i
\end{align*}
which implies the eigenvalues and eigenfunctions of the Laplace-Beltrami operator coincide  with those of the limit of the modified averaging operator. 
\subsection{Vector Diffusion Maps (2011, A. Singer and H.-T. Wu)}
Let the data $\{x_i\}_{i=1}^N$ be i.i.d. with respect to a uniformly bounded probability density function $p(x)$ supported on $M$, i.e. $0< a \leq p(x) \leq b < \infty$ for any $x \in M$. In this part, we use the convention $\iota(x_i)$ to represent the data $x_i$ in the embedding space. 
\subsubsection{Step 1: Approximation to the Parallel Transport Operator}
The authors of this algorithm collect the basis of the local tangent space via the local PCA and then they align them to approximate the parallel transport operator on manifolds. \\\\
Recall Section 2.2 of Chapter 2. If we pick the parameter $\epsilon_{\text{PCA}} = O(N^{-\frac{2}{d+2}})$ of the method of the local PCA, we consider $x_i, x_j \not\in M_{\sqrt{\epsilon_{\text{PCA}}}}=\{x\in M: \min_{y\in \partial M}d_g(x,y) \leq {\sqrt{\epsilon_{\text{PCA}}}}$ but $d_g(x_i, x_j)=O(\sqrt{\epsilon})$. Then the $d\times d$ orthogonal transformation $O_{ij}(:=U V^T$, with the SVD of $O_i^T O_j=U \Sigma V^T$) will approximate$P_{x_i, x_j}$ in the following sense: for any $X\in C^3(TM)$ 
\begin{align*}
O_{ij}\left(\langle \iota_*X(x_j), u_l(x_i) \rangle \right)_{l=1}^d = \left(\langle \iota_*P_{x_i, x_j} X(x_j), u_l(x_i) \rangle \right)_{l=1}^d + O\left(\epsilon_{\text{PCA}}^{\frac{3}{2}}+\epsilon^{\frac{3}{2}}\right),
\end{align*}
where $\{u_l(x_i)\}_{l=1}^d$ is an orthonormal basis determined by the local PCA. For $x_i, x_j \in M_{\sqrt{\epsilon_{\text{PCA}}}}$, we have that for any $X\in C^3(TM)$, 
\begin{align*}
O_{ij}\left(\langle \iota_*X(x_j), u_l(x_i) \rangle \right)_{l=1}^d = \left(\langle \iota_*P_{x_i, x_j} X(x_j), u_l(x_i) \rangle \right)_{l=1}^d + O\left(\epsilon_{\text{PCA}}^{\frac{1}{2}}+\epsilon^{\frac{3}{2}}\right).
\end{align*}
\subsubsection{Step 2: Normalized Kernel and Normalized Connection-Laplacian of the Graph} 
The following steps are as a routine as in Diffusion Map. Firstly, we need to introduce the normalized kernel. Given the local information of a submanifold, define $K_\epsilon (x_i, x_j)$ by
\begin{align*}
       K_\epsilon (x_i, x_j) := K \left( \frac{||\iota(x_i)-\iota(x_j)||_{\mathbb{R}^D}}{\sqrt{\epsilon}}  \right),
\end{align*}
where $||\iota(x_i)-\iota(x_j)||_{\mathbb{R}^D}<\sqrt{\epsilon}$. Then, we define an estimated probability density distribution by
\begin{align*}
          p_\epsilon(x_i):=\sum_{j=1}^N K_\epsilon(x_i, x_j) 
\end{align*}
and the $\alpha$-normalized kernel $K_{\epsilon, \alpha}$ $(0\leq \alpha \leq 1)$ by 
\begin{align*}
      K_{\epsilon, \alpha}(x_i, x_j) := \frac{K_\epsilon(x_i, x_j)}{p_\epsilon^\alpha(x_i)\cdot p_\epsilon^\alpha(x_j)}.
\end{align*}
\\\\
Denote the averaging operator for the vector fields, for a fixed $\epsilon$ by 
\begin{align*}
[D_{\epsilon,\alpha}^{-1}S_{\epsilon, \alpha}](\bar{{X}_i}):= \frac{\sum_{j=1}^N K_{\epsilon,\alpha}(x_i, x_j)O_{ij}\bar{X}_j}{\sum_{j=1}^N K_{\epsilon, \alpha}(x_i, x_j)},
\end{align*}
which can be regarded as the transportation of the vector fields from $T_{x_j}M$ to $T_{x_i}M$ and then averaging them at $T_{x_i}M$. Thus, the \textbf{normalized connection-Laplacian} $[D_{\epsilon,\alpha}^{-1}S_{\epsilon, \alpha}-I]$ on the data graph is defined by
\begin{align*}
    [D_{\epsilon,\alpha}^{-1}S_{\epsilon, \alpha}-I](\bar{{X}_i})= \frac{\sum_{j=1}^N K_{\epsilon,\alpha}(x_i, x_j)O_{ij}\bar{X}_j}{\sum_{j=1}^N K_{\epsilon, \alpha}(x_i, x_j)}-\bar{{X}_i}.
\end{align*}
Recall the normalized Laplacian is formally denoted as
\begin{align*}
L_{\text{RW}}:=D^{-1}L=D^{-1}(D-W)=I-D^{-1}W, 
\end{align*} 
where $D$ and $W$ is the diagonal and weight matrix on a graph respectively. 
\subsubsection{Step 3: Approximation to the Heat Kernel of the Connection-Laplacian} 
Define the averaging operator $\mathcal{A}_{\epsilon, \alpha}$ as
\begin{align*}
    [\mathcal{A}_{\epsilon, \alpha}X](x):= \frac{\int_{M(y)} K_{\epsilon, \alpha}(x,y)P_{x, y}X(y)}{\int_{M(y)} K_{\epsilon, \alpha}(x,y)},
\end{align*}
which has an asymptotic expansion: for $X\in C^3(TM)$ and $x \not\in M_{\sqrt{\epsilon_{\text{PCA}}}}$, then
\begin{align*}
     [\mathcal{A}_{\epsilon, \alpha}X](x) = X(x) + \frac{\epsilon \cdot m_2}{2d \cdot m_0}\left\{ \Delta X(x) + d\cdot \frac{\int_{S^{d-1}}\nabla_\theta X(x)\cdot \nabla_\theta(p^{1-\alpha})(x)d\theta}{p^{1-\alpha}(x)}  \right\} +O(\epsilon^2),
\end{align*}
where  $m_l = \int_{\mathbb{R}^d} ||x||^l \cdot K(||x||)dx$. \\\\
In particular, 
\begin{align*}
  [\mathcal{A}_{\epsilon, 1}X](x) = X(x) + \frac{\epsilon \cdot m_2}{2d \cdot m_0}\Delta X(x) +O(\epsilon^2).
\end{align*}
The asymptotic expansion of the averaging operator $\mathcal{A}_{\epsilon, \alpha}$ contains the connection-Laplacian and potential term, and when $\alpha=1$, the potential term vanishes. \\\\
The theorem in \cite[Theorem 5.3]{singer2012vector} implies that the averaging operator approximates the heat kernel $e^{-t\Delta}$ in $L^2(M)$, i.e.
\begin{align*}
     \lim_{\epsilon \rightarrow 0^+} \mathcal{A}_{\epsilon, 1}^{\frac{t}{\epsilon}} = e^{-t\Delta}.
\end{align*}
\subsubsection{Step 4: Computation via the Connection-Laplacian Operator of Graph}
In this step, we will prove the matrix $D_\alpha^{-1}S_\alpha-I$ where $0\leq \alpha \leq 1$ converges to the connection-Laplacian operator and the potential term. In particular, $D_1^{-1} S_1-I$ converges to the connection-Laplacian operator called the connection-Laplacian on a graph in the literature. According to the work of this step, we can compute the $i$\textsuperscript{th} eigenvector field of $D_1^{-1} S_1-I$, which is a discrete approximation of the $i$\textsuperscript{th} eigenvector field of the connection-Laplacian $\Delta$ over $M$. \\\\
The following equations describe the relations between the normalized connection-Laplacian on a graph and averaging operator.
\begin{itemize}
\item For $x_i \not\in M_{\sqrt{\epsilon_{\text{PCA}}}}$, we have
\begin{align}
[D_{\epsilon,\alpha}^{-1}S_{\epsilon, \alpha}-I](\bar{{X}_i})= \left(\langle \iota_*[\mathcal{A}_{\epsilon, \alpha}X](x_i) \cdot X(x_j), u_l(x_i) \rangle \right)_{l=1}^d + O(\frac{1}{N^{\frac{1}{2}}\epsilon^{\frac{d}{4}-\frac{1}{2}}}+\epsilon_{\text{PCA}}^{\frac{3}{2}}+\epsilon^{\frac{3}{2}}),
\end{align}
where $\bar{X}_i:=\left(\langle \iota_* X(x_i), u_l(x_i) \rangle \right)_{l=1}^d \in \mathbb{R}^d$.
\item For $x_i \in M_{\sqrt{\epsilon_{\text{PCA}}}}$, we have 
\begin{align}
[D_{\epsilon,\alpha}^{-1}S_{\epsilon, \alpha}-I](\bar{{X}_i}) = \left(\langle \iota_* [\mathcal{A}_{\epsilon, \alpha}X](x_i) \cdot X(x_j), u_l(x_i) \rangle \right)_{l=1}^d + O(\frac{1}{N^{\frac{1}{2}}\epsilon^{\frac{d}{4}-\frac{1}{2}}}+\epsilon_{\text{PCA}}^{\frac{1}{2}}+\epsilon^{\frac{3}{2}}),
\end{align}
where $\bar{X}_i:=\left(\langle \iota_* X(x_i), u_l(x_i) \rangle \right)_{l=1}^d \in \mathbb{R}^d$.
\end{itemize}
According to the operator relation (3.3.3) \& (3.3.4) and geometric relations in Section 3.1, we have the following structural results if the manifold has no boundary. 
\begin{itemize}
\item For $\epsilon = O(N^\frac{2}{d+2})$ and $X \in C^3(TM)$, with high probability 
\begin{align*}
\nonumber
 &\frac{1}{\epsilon}[D_{\epsilon,\alpha}^{-1}S_{\epsilon, \alpha}-I](\bar{{X}_i})\\\nonumber
= & \frac{m_2}{2d\cdot m_0} \left( \langle \iota_* \left\{ \Delta X(x_i) + d\cdot \frac{\int_{S^{d-1}}\nabla_\theta X(x_i)\cdot \nabla_\theta(p^{1-\alpha})(x_i)d\theta}{p^{1-\alpha}(x_i)} \right\}, e_l(x_i) \rangle  \right)_{l=1}^d   \\ 
&+ O(\epsilon^{\frac{1}{2}}+N^{-\frac{3}{d+2}} \epsilon^{-1}+N^{-\frac{1}{2}}\epsilon^{-\frac{d}{4}+\frac{1}{2}}),
\end{align*}
where $\{e_l(x_i)\}_{l=1}^d$ is an orthonormal basis for $\iota_*T_{x_i}M$. \\\\
In particular, when $\alpha=1$
\begin{align*}
 &\frac{1}{\epsilon}[D_{\epsilon,1}^{-1}S_{\epsilon,1}-I](\bar{{X}_i})\\
=& \frac{m_2}{2d\cdot m_0} \left( \langle \iota_* \Delta X(x_i), e_l(x_i) \rangle  \right)_{l=1}^d + O(\epsilon^{\frac{1}{2}}+ N^{-\frac{3}{d+2}}\epsilon^{-1}+N^{-\frac{1}{2}}\epsilon^{-\frac{d}{4}+\frac{1}{2}}).
\end{align*}
\item For $\epsilon = O(N^\frac{2}{d+4})$ and $X \in C^3(TM)$, almost surely 
\begin{align*}
\lim_{N\rightarrow \infty} 
\frac{1}{\epsilon}&[D_{\epsilon,\alpha}^{-1}S_{\epsilon, \alpha}-I](\bar{{X}_i})\\
=& \frac{m_2}{2d\cdot m_0}\left( \langle \iota_* \left\{ \Delta X(x_i) + d\cdot \frac{\int_{S^{d-1}}\nabla_\theta X(x_i)\cdot \nabla_\theta(p^{1-\alpha})(x_i)d\theta}{p^{1-\alpha}(x_i)} \right\}, e_l(x_i) \rangle  \right)_{l=1}^d.
\end{align*}
In particular, when $\alpha=1$, we have
\begin{align*}
\lim_{N \rightarrow \infty} \frac{1}{\epsilon}[D_{\epsilon,1}^{-1}S_{\epsilon, 1}-I](\bar{{X}_i}) = \frac{m_2}{2d\cdot m_0} \left( \langle \iota_* \Delta X(x_i), e_l(x_i) \rangle  \right)_{l=1}^d. 
\end{align*}
\end{itemize}
When the manifold has boundary, we have for $x_i \in M_{\sqrt{\epsilon}}$
\begin{align*}
&[D_{\epsilon,\alpha}^{-1}S_{\epsilon, \alpha}](\bar{{X}_i})\\
=&\left( \langle \iota_* P_{x_i, x_0} \left( X(x_0)+\frac{m_1^\epsilon}{m_0^\epsilon}\nabla_{\partial_r} X(x_0) \right), e_l(x_i) \rangle \right)_{l=1}^d +O(\epsilon+N^{-\frac{3}{2(d+1)}}+N^{-\frac{1}{2}}\epsilon^{-\frac{d}{4}-\frac{1}{2}}).
\end{align*}
where $x_0 =\argmin_{y\in\partial M}d(x_i,y)$, constant $m_1^\epsilon=O(\epsilon^{\frac{d}{2}})$ and constant $m_0^\epsilon=O(\epsilon^{\frac{d}{2}+\frac{1}{2}})$. The formal definition of  $m_1^\epsilon$ and $m_0^\epsilon$ are in \cite[B 6]{singer2012vector}. The value $\partial_r$ is the normal direction to the boundary at $x_0$.

\subsection{Framework under the Weighted Laplacian (2005, M. Hein, J-Y Audibert and U.V. Luxburg)}
Observe that the bias term contains the potential term in the asymptotic expansion of the averaging operator. Thus Hein, Audibert and Luxburg consider a generalized version of the convergence via modifying the Laplacian to $\mu$-Laplacian $\Delta_\mu$. In their framework, they show the convergence of the averaging operator to the $\mu$-Laplacian.\\\\
Let $\mu$ be a measure on $M$ defined by 
\begin{align*}
   d\mu := h^2 \omega_g,
\end{align*}
where $h$ is a smooth positive function on $M$. Then, the Laplace operator $\Delta_\mu$ of $(M, \mathcal{B}(M), \omega_g)$ is
\begin{align*}
       \Delta_\mu:=\text{div}_\mu \circ \nabla = \frac{1}{h^2} \text{div}(h^2 \nabla) =\Delta + 2\frac{\langle \nabla h, \nabla \rangle}{h}. 
\end{align*}\\
Since the $\mu$-Laplacian has all the properties as the Laplacian on manifolds (see \cite{grigor2006heat} for details), we could also define the averaging operator $\mathcal{A}_{\epsilon, \mu}$ approaching to the $\mu$-Laplacian $ \Delta_\mu$ when the parameter $\epsilon$ approaches $0^+$. Despite the generalized version of the Laplacian, the main achievement of their approach is that they reduce the bias and variance term simultaneously. For more details, please refer to \cite{hein2005graphs}.


\chapter{Summary}

In this survey, we talk about the some of the most popular techniques used today for nonlinear data reduction algorithms including ISOMAP, LLE, EigenMap, Hessian EigenMap, MVU, LTSA, Diffusion Maps and Vector Diffusion Maps. These algorithms are also called Manifold learning algorithms, since we assume the high dimensional data are located on an embedding submanifolds in higher dimensional Euclidean space and then develop the learning theory in this case. 

In Chapter two, we consider the features from these eight algorithms. We group them as geometric properties and topological properties. Firstly, we get the local information by the local linearity method and the local PCA method. Then, we patch all the local information to a global version. Each algorithm is designed for a special merit of the dimensionality reduction. In the last section, we compare each algorithm via the numerical experiments in the following aspects: computational speed, geometry, noise, distribution, clustering, high dimension, and parameter sensitivity. 

In Chapter three, we discuss the convergence issue including the algorithm EigenMap, Diffusion Maps and Vector Diffusion Map. We talk about the embedding relation in Section 3.1 and reformulate EigenMap, Diffusion Map and Vector Diffusion Map in the language of Operator theory. Section 3.3 talks about the convergence issue of these three algorithms. 

In light of the discussion in these notes, we provide a roughly procedure to acquire a good lower dimensional representation for the higher dimensional data. PCA should be regarded as the most efficient way since it has lowest computation complexity. It takes only seconds to run and can be performed in space with high dimensionality. However, for special data sets with some geometric structure, we should use the true manifold learning algorithms. It's still a question which the best algorithm are when we apply them to the special data set. There are several unknown parameters in the algorithms and there is no criteria for the procedure of picking parameters. For example, for several algorithms, how to guarantees the convergence issue to a solution in the limit of a large data set. In general, we have to run the algorithms one by one and pick the parameters by the experience and randomly in some sense. After the last algorithm (Vector Diffusion Maps) which is purposed in 2011, several new Manifold leaning algorithms came out, for example, t-SNE, Auto-encoder, XOM and etc. The new algorithms performs well in some cases but the major challenges still remain. Since of these challenges, this field or research are still active and charming for the new generation of mathematicians.

\appendix

\chapter*{APPENDICES}
\addcontentsline{toc}{chapter}{APPENDICES}
\chapter{Basic Matrix Analysis}
\label{AppendixA}
The Stiefel manifold $\mathcal{V}_d(\mathbb{C}^D)$ is a set of all orthogonal $d$-frames in $\mathbb{C}^D$, i.e.
\begin{align*}
          \mathcal{V}_d(\mathbb{C}^D) := \{V \in \mathbb{C}^{D\times d}: {V^*}V=I_d \}.
\end{align*}
The topology of $\mathcal{V}_d(\mathbb{C}^D)$ is the subspace topology inherited from $\mathbb{C}^{D\times d}$. Then$\mathcal{V}_d(\mathbb{C}^D)$ is a compact manifold with dimension $2Dd-d^2$.
\section{Courant-Fischer Principle}
The eigenvalue of a Hermitian matrix can be characterized by the Courant-Fischer principle as follows. 
\begin{mdframed}[backgroundcolor=blue!5]
\vspace{0.1cm}
\textbf{Courant-Fischer Principle \cite{reed1978iv}}:
Given a $D\times D$ Hermitian matrix $A$ with the eigenvalues sorted in descending order $\lambda_1 \geq \lambda_2 \geq \dots \geq \lambda_D$, its eigenvalues satisfy
\begin{align*}
\lambda_k = \max_{V \in \mathcal{V}_k(\mathbb{C}^D)} \lambda_{\text{min}} (V^* A V) = \min_{V \in \mathcal{V}_{D-k+1}(\mathbb{C}^D)} \lambda_{\text{max}} (V^* A V).
\end{align*}
\end{mdframed}
\vskip 0.5cm
Note that if matrix $A$ is Hermitian, then all its eigenvalues $\lambda_i$ are real and thus we can compare them with each other. Intuitively, the matrix $V^* A V$ can be regarded as a compression of matrix $A$ in the coordinates spanned by $V$. The optimal value is attained when $V$ is an orthogonal $k-$frame of the subspace generated by the leading $k$ eigenvectors of $A$. 
\section{Singular Value Decomposition}
The singular value decomposition method (SVD) is a common factorization in matrix analysis. It generalizes the eigendecomposition method and benefits of no extra assumptions on the given matrix.  
\begin{mdframed}[backgroundcolor=blue!5] 
\vspace{0.1cm}
\textbf{Singular Value Decomposition \cite{SunMaxA}}:
Suppose $A \in \mathbb{C}^{D\times d}$. Then there exist an unitary matrix $U \in \mathbb{C}^{D\times D}$, an unitary matrix $V \in \mathbb{C}^{d\times d}$ and a diagonal matrix $\Sigma \in \mathbb{R}^{D\times d}$ with the nonnegative entries, such that
\begin{align*}
A= U \Sigma V^*.
\end{align*}
\end{mdframed}
\vskip 0.5cm
Suppose dimension $D\geq d$, we have
\begin{align*}
A&= U \Sigma V^*\\
&=  \begin{pmatrix}
                u_1,\dots,u_D\\
              \end{pmatrix}
                                        \begin{pmatrix}
                                                \sigma_1&          &\\
                                                               &\ddots&\\
                                                                &          &\sigma_d \\
                                                  \hline              
                                                   & 0&\\
                                         \end{pmatrix}
                                         \begin{pmatrix}
                                              v_1^*\\
                                              \vdots\\
                                              v_d^*\\
                                         \end{pmatrix},
\end{align*}
where $\sigma_1 \geq \dots \geq \sigma_d$.\\\\
The diagonal entries of $\Sigma$ are called the singular values of $A$ sorted in descending order. The columns of $U$ and $V$ are called the left-singular vectors and the right-singular vectors of $A$ respectively. It is obvious that the left-singular vectors of $A$ are a set of orthogonal eigenvectors of $AA^*$. And the right-singular vectors of $A$ are a set of orthogonal eigenvectors of $A^*A$. \\\\
In particular, we can express $A$ as a sum of rank-1 matrices, i.e.
\begin{align*}
           A=\sum_{i=1}^d u_i v_i^*.
\end{align*}
One application of the SVD method is the \textbf{low-rank matrix approximation} problem. Let the rank-$k$ approximation $A_k$ be
\begin{align}
           A_k:=\sum_{i=1}^k u_i v_i^*.
\end{align}
Then $A_k$ is spanned by linearly independent vectors $u_1,\dots, u_k$ and thus it has the rank $k$ as its name implies.
\begin{mdframed}[backgroundcolor=blue!5]
\vspace{0.1cm} 
\textbf{Low-rank Matrix Approximation \cite{SunMaxA}}:
Suppose matrix $A\in \mathbb{C}^{D\times d}$ and $k < d \leq D$. Then we have
\begin{align}
\min_{\text{rank}(B)=k} ||A-B||&=\sigma_{k+1}\\
\min_{\text{rank}(B)=k} ||A-B||_F&={\sqrt{\sum_{i=k+1}^d \sigma_i^2}},
\end{align}
where the norm in (A.2.2) is the sup norm and the norm in (A.2.3) is the Frobenius norm. The minimum is attained by $A_k$ defined in (A.2.1).
\end{mdframed}
\section{Ky Fan's Maximum Principle}
The reduction algorithms are always deduced to a kind of eigenvalue problem: 
\begin{mdframed}[backgroundcolor=blue!5] 
\vspace{0.1cm} 
\textbf{Optimization with Orthogonality Constraints \cite{edelman1998geometry}}:
Given a $D\times D$ Hermitian matrix $A$, the optimization problem is
\begin{align*}
\max_{V \in \mathcal{V}_d(\mathbb{C}^D)}  \hspace*{0.5em}  tr\hspace{1pt}(V^* A V).
\end{align*}
\end{mdframed}
\vskip 0.5cm
The optimization problem above can be solved by the optimal matrix $\bar{V}$ consisting of eigenvectors associated to $d$ largest eigenvalue $\lambda_i$ of $A$, i.e.
\begin{align}
\max_{V \in \mathcal{V}_d(\mathbb{C}^D)}  \hspace*{0.5em}  tr\hspace{1pt}(V^* A V) = tr\hspace{1pt}(\bar{V}^* A \bar{V}) = \sum_{i=1}^d \lambda_i.
\end{align}
Formula (A.3.1) are called \textbf{Ky Fan's maximum principle} \cite{fan1950theorem}. It has another version as 
\begin{align*}
\min_{V \in \mathcal{V}_d(\mathbb{C}^D)}  \hspace*{0.5em}  tr\hspace{1pt}(V^* A V) = tr\hspace{1pt}(\bar{V}^* A \bar{V}) = \sum_{i=1}^d \lambda_{D-i+1}.
\end{align*}
In particular, for a fixed $k$ $(1\leq k \leq d \leq D)$, the Ky Fan's $k$-norm of a matrix $A$ is defined as 
\begin{align*}
         ||A||_k :=\sum_{i=1}^k \sigma_i,
\end{align*}
where $\sigma_i$ is the singular values of $A$ sorted in descending order. \\\\
In summary, the optimization problem with orthogonality constraints is solved by computing the largest eigenvalues and associated eigenvectors. For a complete material of matrix analysis, please refer to \cite{SunMaxA}.

\chapter{Linear Manifold Reduction Algorithm}
\label{AppendixB}
A linear manifold is a translation of the subspace of $\mathbb{R}^D$. It is sometimes used as a synonym for the affine subspace or the hyperplane. \\\\
\section{Principle Component Analysis}
The PCA method is proposed by Karl Pearson in 1901 to \emph{present a system of points in plane, three, or higher dimensioned space by the best-fitting straight line or plate} \cite{peason1901lines}. And it has a lot of applications in signal processing, data mining and other fields.\\
\\
In the problem setting, we are given $N$ data points $x_1, \dots, x_N \in \mathbb{R}^D$, and the goal is to find a linear manifold $M$ staying as close to the $N$ data points as possible.
\newpage
\begin{mdframed}[backgroundcolor=blue!5] 
\vskip 0.1cm
\textbf{Principle Component Analysis}: Suppose vectors $\beta_i \in \mathbb{R}^d$ $(i=1,\dots,N)$ satisfy $\sum_{i=1}^N \beta_i =0$. Let the linear manifold $M$ consist of the translation $\mu \in \mathbb{R}^D$ and the $d$-frame $V \in \mathcal{V}_d(\mathbb{R}^D)$. Then $M$ is attained by minimizing the quadratic form
\begin{align}
       \min_{\mu; V; \beta_i} \sum_{i=1}^N  ||x_i-(\mu+V\beta_i)||_2^2.
\end{align}
\end{mdframed}
\vskip 0.5cm
The new representation of data in low dimension is given by $y_k = V^T (x_k-\bar{x})$, where $\bar{x} = \frac{1}{N} \sum_{i=1}^N x_i $ and the $d-$frame $V$ consists of eigenvectors associated with the largest $d$ eigenvalues of the scatter matrix $\Sigma_N$. Let $X:=(x_1, \dots, x_N)$ be a $D\times N$ data matrix and the centering data matrix be $\bar{X} := X(I_N-\frac{1}{N}1\cdot 1^T)$, where $I_N-\frac{1}{N}1\cdot 1^T$ is the centering matrix denoted by $C_N$. Then the scatter matrix $\Sigma_N$ is defined as 
\begin{align*}
       \Sigma_N := \bar{X} \bar{X}^T \overset{(1)}{=} X(I_N-\frac{1}{N}1\cdot 1^T)X^T,
\end{align*}
where vector 1 denotes all one $N$-dimensional vector and equality (1) is derived from the fact $C_N$ is idempotent, i.e. $C_N^k = C_N$, for any $k=1,2,\dots$\\\\
PCA methods holds the max variance property, i.e.
\begin{mdframed}[backgroundcolor=blue!5] 
\vskip 0.1cm
\textbf{Max Variance Property for PCA}: The linear manifold $M$ in (B.1.1) maximizes the variance, i.e.
\begin{align}
       \max_{V \in \mathcal{V}_d(\mathbb{R}^D)} &\sum_{i=1}^N ||y_i - \frac{1}{N}\sum_{j=1}^N y_j||_2^2, \\
\nonumber                                        \text{s.t.    }& y_i = V^Tx_i.
\end{align}
\end{mdframed}
\vskip 0.5cm
If we assume $\sum_{i=1}^N x_i =0$, problem (B.1.1) and problem (B.1.2) are equivalent due to the Pythagorean theorem. In general, we choose dimension $d$ such that for given parameter $\rho$,
\begin{align}
            \frac{\sum_{i=1}^d \lambda_i}{\sum_{i=1}^N \lambda_i} \geq \rho  \hspace*{1em} \text{where  } \lambda_i \in \text{Eig }(\Sigma_N).
\end{align}
\subsubsection{Simple Calculation for PCA}
The following procedure is suggested by \cite{BAsTEN}. Firstly, we differentiate quadratic form (B.1.1) with respect to $\mu$ and since $\sum_{i=1}^N \beta_i =0$ and have
\begin{align*}
    \mu = \bar{x}.
\end{align*}
Because each $\beta_i$ involves only one term of the sum, we then differentiate each $||x_i-(\bar{x}+V\beta_i)||_2^2$ with respect to $\beta_i$ and get
\begin{align*}
    \beta_i = V^T(x_i-\bar{x}).
\end{align*}
Plug each $\beta_i$ into (B.1.1), and we have
\begin{align}
 \nonumber   &\min_{V \in \mathcal{V}_d(\mathbb{R}^D)}  \sum_{i=1}^N ||x_i-\bar{x}-VV^T(x_i-\bar{x})||_2^2\\
  =& \sum_{i=1}^N (x_i-\bar{x})^T(x_i-\bar{x}) - \min_{V \in \mathcal{V}_d(\mathbb{R}^D)}  \sum_{i=1}^N (x_i-\bar{x})^TVV^T(x_i-\bar{x}).                                                                    
\end{align}
The left term of (B.1.4) does not contain the variable $V$, thus we only need to consider the right term
\begin{align}
\nonumber         &\max_{V \in \mathcal{V}_d(\mathbb{R}^D)}  \sum_{i=1}^N (x_i-\bar{x})^TVV^T(x_i-\bar{x})  \\
\nonumber      =&\max_{V \in \mathcal{V}_d(\mathbb{R}^D)} tr(V^T\sum_{i=1}^N (x_i-\bar{x})(x_i-\bar{x})^T V ) \\
            =&\max_{V \in \mathcal{V}_d(\mathbb{R}^D)}  N\cdot tr(V^T\Sigma_NV).                                                               
\end{align}
By Ky Fan's Maximum Principle (A.3.1), the $d-$frame $V$ for (B.1.5) consists of eigenvectors associated with the $d$ largest eigenvalues of the scatter matrix $\Sigma_N$. 
\section{Multidimensional Scaling}
Multidimensional scaling is a common technique putting the distance-like data with dissimilarities between each other into Euclidean space at the same time preserving the dissimilarities. It is proposed by Young and Householder in 1938 in order to recover the coordinates of the cities with the information of distances between each other \cite{young1938discussion}. The problem is described as follows:
\begin{mdframed}[backgroundcolor=blue!5] 
\vskip 0.1cm
\textbf{Great Britain Problem}: We do not know the exact coordinates of $N$ cities, but we know the distances $d_{ij}$ between any pair of cities $x_i\in \mathcal{R}^D$ and $x_j\in \mathcal{R}^D$. Then we can recover the coordinates for the $N$ cities up to isomorphism. 
\end{mdframed}
\vskip 0.5cm
\emph{Up to isomorphism} is in the sense that the coordinates are unique up to translations, rotations and reflections. In practice, the quantity between any two data we collect is not Euclidean. Then we apply multidimensional scaling method to achieve an Euclidean coordinate system. In the new coordinate system, the Euclidean distance between any two data is similar to the original quantity. 
\subsubsection{Simple Calculation for Great Britain Problem}
The calculation procedure is suggested by Cox \cite{cox2000multidimensional}. Let $x_1, \dots, x_N$ be $N$ cities in $\mathbb{R}^N$. We always assume $\sum_{i=1}^N x_i = 0$. Suppose $x_i$'s are all linearly independent. The distance $d_{ij}$ between data $x_i$ and $x_j$ is Euclidean, and we write $d_{ij}$ as
\begin{align}
        d_{ij}^2 = (x_i-x_j)^T\cdot (x_i-x_j)
\end{align} 
Summing (B.2.1) over the index $i$ and index $j$, we have three equalities
\begin{align*}
    \frac{1}{N}\sum_{i=1}^N d_{ij}^2 &= \frac{1}{N}\sum_{i=1}^N x_i^T x_i +x_j^Tx_j,\\
     \frac{1}{N}\sum_{j=1}^N d_{ij}^2 &= \frac{1}{N}\sum_{j=1}^N x_j^T x_j +x_i^Tx_i,\\
      \frac{1}{N^2}\sum_{i,j}^N d_{ij}^2 &= \frac{2}{N}\sum_{i=1}^N x_i^T x_i.
\end{align*}
The Gram matrix $G$ of a data matrix $X=(x_1, \dots, x_N)$ is defined by 
\begin{align}
             G := X^T \cdot X
\end{align}
with the entry $G(i, j)$ as
\begin{align}
 \nonumber    G(i,j)& = x_i^Tx_j\\
 \nonumber           & = -\frac{1}{2}(d_{ij}^2-x_i^Tx_i -x_j^Tx_j)\\
 \nonumber          & = -\frac{1}{2}(d_{ij}^2-\frac{1}{N}\sum_{i=1}^N d_{ij}^2 -\frac{1}{N}\sum_{j=1}^N d_{ij}^2+ \frac{1}{N^2}\sum_{i,j}^N d_{ij}^2)\\
            & = a_{ij}-\frac{1}{N}\sum_{i=1}^N a_{ij}-\frac{1}{N}\sum_{j=1}^N a_{ij}+ \frac{1}{N^2}\sum_{i,j}^N a_{ij},
\end{align} 
where $a_{ij}:=-\frac{1}{2}d_{ij}^2$. \\\\
By (B.2.3), we can write $G$ in a form
\begin{align*}
      G=C_N\cdot A \cdot C_N,
\end{align*}
where $C_N$ is the centering matrix $C_N:=I_N-\frac{1}{N}1\cdot 1^T$ and $A$ is the matrix with entry $a_{ij}$.\\\\
According to (B.2.2), the Gram matrix $G$ is positive semi-definite with rank $N$. Then, we applying the SVD method on the Gram matrix $G$ and get 
\begin{align}
      G=V\cdot \Sigma\cdot V^T,
\end{align}
where $V$ are orthogonal and $\Sigma$ is diagonal with $N$ singular values. \\\\
Comparing (B.2.2) with (B.2.4), we get Euclidean coordinates for data $X:=\Sigma^\frac{1}{2}\cdot V^T$.\\\\\\
In the case the quantities are not Euclidean, we denote the quantity as $\delta_{ij}$ and call them dissimilariities. Observe the computation procedure above. We can still get a Gram matrix $G_\delta$ and matrix $A_\delta$ corresponding to the dissimilarities, i.e.
\begin{align}
                 \delta_{ij} \overset{-\frac{1}{2}\delta_{ij}^2} {\Longrightarrow} A_\delta \overset{C_N A_\delta C_N}{\Longrightarrow} G_\delta \overset{SVD} {\Longrightarrow} X_\delta.
\end{align}
\clearpage
\noindent
The coordinates $X_\delta$ in (B.2.5) may not be Euclidean. We need to come up with a way to make it Euclidean. A method to solve this problem is based on the following observation. 
\begin{mdframed}[backgroundcolor=blue!5] 
\vskip 0.1cm
\textbf{Observation:}
A Gram Matrix $G_\delta$ generated by the procedure (B.2.5) is positive semi-definite $\Leftrightarrow$ Dissimilarities $\delta_{ij}$ are Euclidean, i.e. 
$\delta_{ij}^2 = ||x_i-x_j||_2^2$.
\end{mdframed}
\vskip 0.5cm
One common method for converting a Gram Matrix $G_\delta$ to a positive semi-definite one is to add an appropriate constant $c$ on $\delta$. There are a lot of techniques on how to choose an appropriate constant. After we collect the non-Euclidean dissimilarities $\delta$, run the procedure (B.2.5). If the Gram Matrix $G_\delta$ is not positive semi-definite, we continue to add some constant to $\delta$ until a new Gram Matrix $G_\delta$  is positive semi-definite. Then we decompose the Gram matrix and get the coordinates $X_{\delta'}$ with Euclidean distance between each pair of points. These Euclidean distances are equal to the updated dissimilarities denoted by ${\delta}'$. We use rule (B.1.3) to pick a lower dimension. In summary, this procedure is called \textbf{multidimensional scaling}. For more details, please refer to Cox \cite{cox2000multidimensional}.

\chapter{Laplace-Beltrami Operator and Hessian Tensor on Manifolds}
\label{AppendixC}
\section{Definition}
Given a Riemannian manifold $(M^d, g)$ and a $C^\infty(M)$-function $f$, the \textbf{gradient} operator $\nabla$ is defined as
\begin{align*}
            \nabla: C^\infty(M) \longrightarrow \Gamma(TM),
\end{align*} 
such that 
\begin{align}
 g(\nabla f, X)=df(X) \hspace*{1em} \text{for any }X \in \Gamma(TM).
\end{align}
The \textbf{Hessian tensor} Hess$(f) \in \Gamma(T^*(M) \otimes T^*(M))$ is defined as 
\begin{align*}
        \text{Hess}(f):=\frac{1}{2} \mathcal{L}_{\nabla f} g,
\end{align*}
where $\mathcal{L}$ is the Lie derivative and $\nabla$ is the gradient. \\\\
The \textbf{divergence} operator div is defined as 
\begin{align*}
            \text{div}: \Gamma(TM) \longrightarrow C^\infty(M),
\end{align*} 
such that 
\begin{align}
 d(i_X \omega)=(div X)\omega  \hspace*{1em} \text{for any }X \in \Gamma(TM),
\end{align}
where $\omega=$ $d$ vol$_d:=\sqrt{|g|}\cdot dx_1\wedge\dots\wedge dx_n$.\\\\
The \textbf{Laplace-Beltrami} operator $\Delta$ on $(M,g)$
\begin{align*}
            \Delta: C^\infty(M) \longrightarrow C^\infty(M)
\end{align*} 
is defined as
\begin{align}
 \Delta=-\text{div}\circ \nabla.
\end{align}
\section{In Local Coordinates}
In local coordinates $(U,\varphi_U; x_1, \dots, x_d)$, since $\nabla f$ is a section on the tangent bundle, one can write
\begin{align*}
      \nabla f :=\sum_{i=1}^d \xi^i \frac{\partial}{\partial x_i},
\end{align*}
where $\xi^i$ are $C^\infty(M)$-function.
\begin{itemize}
\item For the gradient part, we pick $X=\frac{\partial}{\partial x_j}(j=1,\dots,d)$. Then  
\begin{align}
                df(\frac{\partial}{\partial x_j})&=\frac{\partial f}{\partial x_j}(:=\frac{\partial f \circ \varphi_U^{-1}}{\partial x_j})\\
                g(\nabla f, \frac{\partial}{\partial x_j}) &= \sum_{i=1}^d \xi^i g(\frac{\partial}{\partial x_i},\frac{\partial}{\partial x_j})
                                                                = \sum_{i=1}^d \xi^i \cdot g_{ij}.
\end{align}
According to the definition (C.1.1), we compare (C.2.1) with (C.2.2) and get
\begin{align*}
   \xi^i = \sum_{j=1}^d g^{ij}\cdot \frac{\partial f}{\partial x_j}.
\end{align*}
Thus, the gradient of $f$ in local coordinates is 
\begin{align*}
    \nabla f = \sum_{i,j}^d g^{ij}\cdot \frac{\partial f}{\partial x_j}\cdot \frac{\partial}{\partial x_i}.
\end{align*}
\item For the Hessian tensor part, we compose the Hessian tensor with the coordinate vector fields and get 
\begin{align}
\nonumber   2\text{Hess}(f)(\frac{\partial}{\partial x_l}, \frac{\partial}{\partial x_k}):=&[\mathcal{L}_{\nabla f} g](\frac{\partial}{\partial x_l}, \frac{\partial}{\partial x_k})\\
 =& D_{\nabla f} g_{lk}- g(\mathcal{L}_{\nabla f}\frac{\partial}{\partial x_l}, \frac{\partial}{\partial x_k})-g( \frac{\partial}{\partial x_l}, \mathcal{L}_{\nabla f}\frac{\partial}{\partial x_k}).
\end{align}
Notice that
\begin{align*}
     D_{\nabla f} g_{lk} &= \sum_{i,j}^d g^{ij} \frac{\partial f}{\partial x_j} \frac{\partial g_{lk}}{\partial x_i}, \\
 \mathcal{L}_{\nabla f}\frac{\partial}{\partial x_l}&= [\nabla f, \frac{\partial}{\partial x_l}]=-\mathcal{L}_\frac{\partial}{\partial x_l} \nabla f. 
\end{align*}
Thus, formula (C.2.3) becomes 
\begin{align*}
&D_{\nabla f} g_{lk}- g(\mathcal{L}_{\nabla f}\frac{\partial}{\partial x_l}, \frac{\partial}{\partial x_k})-g( \frac{\partial}{\partial x_l}, \mathcal{L}_{\nabla f}\frac{\partial}{\partial x_k})\\
=&\sum_{i,j}^d g^{ij} \frac{\partial f}{\partial x_j} \frac{\partial g_{lk}}{\partial x_i} +g(\mathcal{L}_\frac{\partial}{\partial x_l} \nabla f,  \frac{\partial}{\partial x_k})+ g(\frac{\partial}{\partial x_l},  \mathcal{L}_\frac{\partial}{\partial x_k} \nabla f)\\
=&\sum_{i,j}^d g^{ij} \frac{\partial f}{\partial x_j} \frac{\partial g_{lk}}{\partial x_i} +g(\frac{\partial }{\partial x_l}(\sum_{i,j}^d g^{ij} \frac{\partial f}{\partial x_j})\frac{\partial }{\partial x_i}, \frac{\partial }{\partial x_k})+g( \frac{\partial }{\partial x_l},\frac{\partial }{\partial x_k}(\sum_{i,j}^d g^{ij} \frac{\partial f}{\partial x_j})\frac{\partial }{\partial x_i})\\
=&\sum_{i,j}^d g^{ij} \frac{\partial f}{\partial x_j} \frac{\partial g_{lk}}{\partial x_i} + \sum_{i,j}^d(\frac{\partial g^{ij} }{\partial x_l} \frac{\partial f}{\partial x_j} + \frac{\partial^2 f}{\partial x_j \partial x_l}g^{ij})g_{ik}+\sum_{i,j}^d(\frac{\partial g^{ij} }{\partial x_k} \frac{\partial f}{\partial x_j} + \frac{\partial^2 f}{\partial x_j \partial x_k}g^{ij})g_{li}\\
=&\sum_{i,j}^d [2\cdot \frac{\partial^2 f}{\partial x_l \partial x_k}+\frac{\partial f}{\partial x_j}(g^{ij}\frac{\partial g_{lk}}{\partial x_i}-g^{ij}\frac{\partial g_{ik}}{\partial x_l}-g^{ij}\frac{\partial g_{li}}{\partial x_l})]\\
=&\sum_{i,j}^d [2\cdot \frac{\partial^2 f}{\partial x_l \partial x_k}+\frac{\partial f}{\partial x_j}g^{ij}(\frac{\partial g_{lk}}{\partial x_i}- \frac{\partial g_{ik}}{\partial x_l}-\frac{\partial g_{li}}{\partial x_l})]\\
=& 2\sum_{i,j}^d(\frac{\partial^2 f}{\partial x_l \partial x_k}-\Gamma_{kl}^j \frac{\partial f}{\partial x_j}),
\end{align*}
which implies 
\begin{align*}
   \text{Hess}(f)=\sum_{i,j}^d(\frac{\partial^2 f}{\partial x_l \partial x_k}-\Gamma_{kl}^j \frac{\partial f}{\partial x_j}) \cdot dx^l\otimes dx^k,
\end{align*}
where $\Gamma_{kl}^j$ are the Christoffel symbols of the metric $g$.
\item For the Laplace-Beltrami operator part, let $X:=\sum_{i=1}^d \eta^i \frac{\partial}{\partial x_i}$. Then
\begin{align*}
     i_X\omega(\frac{\partial}{\partial x_1}, \dots, \widehat{\frac{\partial}{\partial x_i}}, \dots, \frac{\partial}{\partial x_d})&=
     \omega(X, \frac{\partial}{\partial x_1}, \dots, \widehat{\frac{\partial}{\partial x_i}}, \dots, \frac{\partial}{\partial x_d})\\
     &=(-1)^{i-1} \omega(\frac{\partial}{\partial x_1}, \dots, X, \dots, \frac{\partial}{\partial x_d})\\
     &=(-1)^{i-1}\sqrt{|g|}dx_1\wedge \dots \wedge dx_d(\frac{\partial}{\partial x_1}, \dots, X, \dots, \frac{\partial}{\partial x_d})\\
     &=(-1)^{i-1}\eta^i\sqrt{|g|}.
\end{align*}
Thus, 
\begin{align*}
      i_X \omega=\sum_{i=1}^d (-1)^{i-1}\eta^i\sqrt{|g|}  \cdot dx_1\wedge \dots \wedge \widehat{dx_i}\wedge \dots \wedge dx_d.
\end{align*}
Take the exterior derivative of $i_X\omega$
\begin{align*}
    d(i_X\omega) &=\sum_{i=1}^d (-1)^{i-1}\eta^i\sqrt{|g|} \cdot dx_1\wedge \dots \wedge \widehat{dx_i}\wedge \dots \wedge dx_d\\
    &=\sum_{i=1}^d  \frac{\partial}{\partial x_i}(\eta^i\sqrt{|g|}) \cdot dx_1\wedge \dots \wedge dx_d \\
    &=\frac{1}{\sqrt{|g|}} \sum_{i=1}^d \frac{\partial}{\partial x_i}(\eta^i\sqrt{|g|}) \cdot \omega.
\end{align*}
By definition (C.1.2), we get the divergence operator in local coordinates 
\begin{align*}
     \text{div} X =\frac{1}{\sqrt{|g|}} \sum_{i=1}^d \frac{\partial}{\partial x_i}(\eta^i\sqrt{|g|}).
\end{align*}
By definition (C.1.3), we get the Laplace-Beltrami operator in local coordinates
\begin{align*}
       \Delta =-\frac{1}{\sqrt{|g|}} \sum_{i=1}^d \frac{\partial}{\partial x_i}(g^{ij}\sqrt{|g|}\frac{\partial}{\partial x_j}).
\end{align*}
\end{itemize}
\subsubsection{Note} 
\begin{itemize}
\item In most of the manifold learning methods, we only require $f$ to be $C^2(M)$. Then $\nabla f$ and $\Delta f$ are as well defined as $C^\infty(M)$.
\item In the geodesic normal coordinates, the Laplace-Beltrami operator is 
         \begin{align*}
                   \Delta=-\sum_{i=1}^d \frac{\partial^2}{\partial x_i^2}.
         \end{align*}
\item \textbf{Green's Formula}: If $f_1$ and $f_2$ are $C^\infty(M)$ function with a compact support on $(M,g)$, then
         \begin{align*}
               \int_{M} (\Delta f_1)f_2 \omega = \int_{M} (\Delta f_2)f_1 \omega =-\int_M g(\nabla f_1, \nabla f_2)\omega.
         \end{align*} 
\end{itemize}
For a complete material of differential geometry, please refer to Chern \cite{chern1950lecture}, Petersen \cite{petersen2006riemannian} or Jost \cite{jost2008riemannian}.

\chapter{Heat Operator on Manifolds}
\label{AppendixD}
In this section, we assume manifold $M$ is compact without boundary. 
\section{Sobolev Space}
In this section, we assume  $p \in [1,\infty)$.\\\\
A \textbf{multi-index} $\alpha=(\alpha_1,\dots,\alpha_d)$ is an index for the multiple partial differentiation index 
\begin{align*}
             D^\alpha :=(\frac{\partial}{\partial x_1})^{\alpha_1}(\frac{\partial}{\partial x_2})^{\alpha_2}\dots (\frac{\partial}{\partial x_d})^{\alpha_d}.
\end{align*}
We usually denote $|\alpha|:= \sum_{i=1}^d \alpha_i$ as the sum of indices, where all $\alpha_i$ are nonnegative integers. \\\\
Given an open subset $\Omega \subset \mathbb{R}^d$, a multi-index $\alpha=(\alpha_1,\dots,\alpha_d)$ and for any $f \in \mathcal{L}_{\text{loc}}^1(\Omega)$, if there exists a locally integrable function $g\in \mathcal{L}_{\text{loc}}^1(\Omega)$ satisfying 
\begin{align*}
       \int_\Omega f D^\alpha \phi = (-1)^{|\alpha|} \int_\Omega g \phi \hspace*{1em} \text{for any test function } \varphi \in C_0^\infty(\Omega),
\end{align*}
we say $g$ is the \textbf{weak derivative} of $f$ with the multi-index $\alpha$. We denote $g$ as $D^\alpha f$.\\\\
\subsubsection{Note}
\begin{itemize}
\item The weak derivative for a locally integral function may not exist, but it'll be unique if it exists.
\item The set of test functions $\mathcal{D}(\Omega)$ is dense in $L^p(\Omega)$ space.
\end{itemize}
The topology of $\mathcal{D}(\Omega)$ is defined by the convergence for sequences. Say $\varphi_i$ \textbf{converges} to $\varphi$, if there exists a compact set $K\subset \Omega$ such that supp $\varphi_n \subset K$ for every $n \in \mathbb{N}$ and $D^\alpha \varphi_n$ uniformly converges to $D^\alpha \varphi$ on $\Omega$ for any multi-index $\alpha$ as $n\rightarrow \infty$.\\\\
The \textbf{distribution} $\mathcal{D}'(\Omega)$ is a set of all continuous functionals on $\mathcal{D}(\Omega)$, i.e.
\begin{align*}
 \mathcal{D}'(\Omega) :=\{T:\mathcal{D}(\Omega)\rightarrow \mathbb{R}|T(\varphi_n)\overset{n\rightarrow \infty}{\longrightarrow} T(\varphi) \text{, for any } \varphi_n \in \mathcal{D}\overset{n\rightarrow \infty}{\longrightarrow} \varphi \in \mathcal{D} \}.
\end{align*}
\subsubsection{Note}
\begin{itemize}
 \item In particular, for any $f\in\mathcal{L}_{\text{loc}}^p(\Omega)$, we can define a regular distribution 
          \begin{align*}
                          T_f(\varphi):=\int_\Omega f \varphi \hspace*{1em}\text{ for any } \varphi \in \mathcal{D}(\Omega).
          \end{align*}
\item The $\alpha$-derivative of distribution $T$ is defined as
          \begin{align*}
                D^\alpha T(\varphi) = (-1)^{|\alpha|} T(D^\alpha \varphi) \hspace*{1em}\text{ for any } \varphi \in \mathcal{D}(\Omega).
          \end{align*}
\end{itemize}
Suppose a subset $\Omega \in \mathbb{R}^d$ is open and connected with a compact closure $\bar{\Omega}  \in \mathbb{R}^d$. According to (D.1.5), the $\alpha$-derivative of $T_f$ exists for any index $\alpha$. Thus, the \textbf{Sobolev space} $W^{k,p}(\Omega)$ is defined as
\begin{align*}
     W^{k,p}(\Omega) =\{u\in \mathcal{L}_{\text{loc}}^p(\Omega) | D^\alpha(u)\in\mathcal{L}_{\text{loc}}^p(\Omega) \hspace*{1em} \forall \hspace*{0.3em}\alpha, \hspace*{0.3em} |\alpha|\leq k\}
\end{align*}
with the norm
\begin{align*}
   ||u||_{k,p} :=(\int_\Omega \sum_{|\alpha|\leq k}|D^\alpha u|^p)^{\frac{1}{p}}.
\end{align*}
\subsubsection{Note}
\begin{itemize}
\item $W^{k,p}(\Omega)$ is a Banach space with respect to the $||\cdot||_{k,p}$ norm.
\item A subset $S(\Omega)$ of $C^k(\Omega)$ is defined as
\begin{align*}
S(\Omega):=\{u\in C^k(\Omega)| \hspace*{0.3em}||u||_{k,p}<\infty\}.
\end{align*}
Since the subsect $S(\Omega)$ is not always a complete metric space, we consider the completion of $S(\Omega)$ denoted as $H^{k,p}(\Omega)$. By the Meyers-Serrin theorem \cite{gilbarg2015elliptic}, we have
\begin{align*}
   H^{k,p}(\Omega) = W^{k,p}(\Omega).
\end{align*}
\item Let $W_0^{k,p}(\Omega)$ be the closure of $C_0^k(\Omega)$ in $W^{k,p}(\Omega)$ with respect to the $||\cdot||_{k,p}$ norm. Then $W_0^{k,p}(\Omega)$ is a closed subspace of $W^{k,p}(\Omega)$ and thus it is also a Banach space. In particular, if $\Omega$ is the whole space $\mathbb{R}^d$, we have
\begin{align*}
       W_0^{k,p}(\mathbb{R}^d) =W^{k,p}(\mathbb{R}^d).
\end{align*}
\item $W^{k,2}(\Omega)$ is called the \textbf{Hilbert-Sobolev} space and denote as
\begin{align*}
       H^k(\Omega):=W^{k,2}(\Omega)
\end{align*}
with the norm 
\begin{align*}
||u||_{H^k(\Omega)}:= (\int_\Omega \sum_{|\alpha|\leq k} |D^\alpha u |^2)^{\frac{1}{2}}.
\end{align*}
It is a Hilbert space with the inner product 
\begin{align*}
 \langle u,v\rangle_{H^k(\Omega)} := \int_\Omega \sum_{|\alpha|\leq k} D^\alpha u D^\alpha v.
\end{align*}
The Hilbert-Sobolev space $H^k(\Omega)$ is related to Fourier transform theory on $L^2(\Omega)$ by the following equality
\begin{align*}
H^k(\Omega)=\{ u\in \mathcal{L}_{\text{loc}}^2(\Omega) | (\int_\Omega \sum_{|\alpha|\leq k} (1+|\xi|^2)^k|\hat{u}(\xi)|^2)^{\frac{1}{2}} <\infty \},
\end{align*}
where $\hat{u}(\xi)$ is the Fourier transform of $u$.
\item \textbf{Sobolev Embedding Theorem}: Let $\Omega \subset \mathbb{R}^d$ be an open subset and $k,m \in \mathbb{N}_0$. If $k>m+\frac{d}{2}$, then for any $f \in H^k(\Omega)$, there exists $g\in C^m(\Omega)$ satisfying $f=g$ a.e.
\item \textbf{Poincare Inequality} (Simple Case): Let $\Omega \subset \mathbb{R}^d$ be an open bounded subset. Then there exists a constant $C:=C(p, \Omega)$ satisfying 
\begin{align}
||u||_{L^p(\Omega)}\leq C||Du||_{L^p(\Omega)} \hspace*{1em} \text{ for all } u\in W_0^{1,p}(\Omega).
\end{align}
\item \textbf{Sobolev Space on manifolds}: Pick a locally finite coordinate covering $\{(U_\alpha, \phi_\alpha)\}_\alpha$ of $M$. Suppose $h_\alpha$ is the corresponding partition of unity so that supp($h_\alpha) \in U_\alpha$, Sobolev space on manifolds $H^k(M)$ is defined as the completion of $C_0^\infty(M)$ with the norm 
\begin{align*}
      ||u||_{H^k(M)}:=(\sum_\alpha ||(h_\alpha u)\circ \phi_\alpha^{-1}||_{H^k(\phi_\alpha(U_\alpha))})^{\frac{1}{2}}.
\end{align*}
\end{itemize}
\section{Homogeneous Heat Equation and Fundamental Solutions}
Suppose $u(x,t): M \times (0,\infty) \rightarrow \mathbb{R}$ be a $C^{2,1}(M \times (0,\infty) )$ function. The homogeneous heat equation is given by 
\begin{align}
   (\Delta+\partial_t )u(x,t)&=0 \hspace*{1em} (x,t)\in M\times(0,\infty)\\
   u(x,0)&=f(x) \hspace*{1.2em} x\in M.
\end{align}
The fundamental solution $u_y(x,t)$ for the homogeneous heat equation (D.2.1) and (D.2.2) at point $y \in M$ satisfying
\begin{align}
\nonumber   (\Delta+\partial_t )u_y(x,t) &= 0 \hspace*{0.5em} \text{ for all } t>0\\
         \lim_{t\rightarrow{0^+}}u_y(x,t)&=\delta_y(x),
\end{align}
where $\delta_y(x)$ is the Dirac distribution centered at $y$.\\\\
\subsubsection{Note}
\begin{itemize}
\item The solution to the homogeneous heat equation (D.2.1) and (D.2.2) is unique. 
\item The limit in (D.2.3) is in the distribution sense, i.e.
         \begin{align*}
                   \lim_{t\rightarrow{0^+}} \int_M u_y(x,t)\varphi(x) = \varphi(x) \hspace*{0.5em} \text{ for all }\varphi \in C_0^\infty(M) \text{ and any } x \in M.
         \end{align*}
\item For the heat equation defined on $\mathbb{R}^d$, the fundamental solution is
         \begin{align*}
                    u_y(x,t):=\frac{1}{(4\pi t)^{\frac{d}{2}}} e^{-\frac{||y-x||^2}{4t}}.
         \end{align*}
         However, it is not trivial to prove the existence of the fundamental solution on compact manifolds. Please refer to \cite[Section 6.5]{canzani2013analysis} for more details.  
\item The general heat equation on manifolds is defined as
\begin{align}
   (\Delta+\partial_t )u(x,t)&=F(x,t) \hspace*{1em} (x,t)\in M\times(0,\infty)\\
   u(x,0)&=f(x) \hspace*{1.2em} x\in M.
\end{align}
By the Duhamel's principle, the solution to the general heat equation (D.2.4) and (D.2.5) is given by
\begin{align*}
        \bar{u}(x,t):=\int_{M(y)} u_y(x,t)f(y) + \int_0^t \int_{M(y)} u_y(x,m) F(y,t-m),
\end{align*}
where $f\in C(M)$ and $F(x,t) \in C(M\times(0,\infty))$.
\item In particular, we have 
\begin{align*}
\int_{M(y)} u_y(x,t)=1 \hspace*{0.5em} \text{ for all } x\in M \text{ and  }t\in(0,\infty).
\end{align*}
\end{itemize}
\section{Heat Semigroup}
We call the heat semigroup $e^{-t\Delta}:L^2(M)\rightarrow L^2(M)$ as
\begin{align*}
e^{-t\Delta}f(y):=\int_{M(y)} u_y(x,t) f(y).
\end{align*}
\subsubsection{Note}
\begin{itemize}
\item Heat semigroup $e^{-t\Delta}$ is self-adjoint, positive and compact. 
\item $e^{-t\Delta}f$ converges uniformly to a harmonic function as $t\rightarrow \infty$ for any $f\in L^2(M)$. It converges to a constant function if $\partial M=\varnothing$.
\item \textbf{Sturm-Liouville Decompostion}: There exists a complete orthonormal basis $\{\varphi_i\}_{i=1}^\infty$ of $L^2(M)$, where $\varphi_i$ is the eigenfunctions of $\Delta$ corresponding to the eigenvalues $\lambda_i$ in ascending order. Then $\varphi_i\in C^\infty(M)$ and  
\begin{align*}
u_y(x,t)=\sum_{j=1}^\infty e^{-\lambda_i t}\varphi_j(x)\varphi_j(y).
\end{align*}
\end{itemize}
\section{Eigenvalue Problem on the Riemannian Manifolds}
Suppose the eigenvalues of $\Delta$ on $M$ are $\lambda_1 (=0) \leq \lambda_2 \leq \dots$ and the corresponding $L^2$-normalized eigenfunctions are $\varphi_1 \leq \varphi_2 \leq \dots$.\\\\
Let the space $H_1^2(M)$ be the completion of $C^\infty(M)$ with respect to the norm
\begin{align*}
      ||\varphi||_{H_1^2(M)}^2:=\int_M \varphi^2 +\int_M |\nabla \varphi|^2.
\end{align*}
Then $H_1^2(M)=\{f\in L^2(M): \sum_{i=0}^\infty \lambda_i \langle f, \varphi_i \rangle^2<\infty \}$.\\\\
Now we define a bilinear form $B: C^\infty(M)\times C^\infty(M)\rightarrow \mathbb{R}$ by
\begin{align*}
      B(f,h):=\int_M \nabla f\cdot \nabla h.
\end{align*}
Then the bilinear form $B$ can be naturally extended to $\bar{B}:H_1^2(M)\times H_1^2(M)  \rightarrow \mathbb{R}$
\begin{align*}
          \bar{B}(f,h):=\lim_{i\rightarrow \infty} B(f_i,h_i),
\end{align*}
where $f_i$ and $h_i$ approach to $f$ and $h$ in $H_1^2(M)$ respectively, as $i\rightarrow \infty$.
\clearpage
\noindent
Denote 
\begin{align*}
 E_k:=\{\varphi_1,\dots,\varphi_{k-1}\}^\perp.
\end{align*}
Let $\mathcal{V}_k$ be a collection of all subspaces $V\in C^\infty(M)$ of dimension k. Then we have 
\begin{align*}
     \lambda_k=\inf_{\phi \in H_1^2(M) \cap E_k} \frac{D(\phi,\phi)}{\int_M |\phi|^2}.
\end{align*}
The Courant-Fisher theorem on manifolds is 
\begin{align*}
\lambda_k&=\sup_{V\in\mathcal{V}_{k-1}}  \inf_{\phi \in (V^\perp \cap H_1^2(M))/\{0\}} \frac{D(\phi,\phi)}{\int_M |\phi|^2} \\
                 &= \inf_{V\in\mathcal{V}_k}  \hspace{0.9em}  \sup_{\phi \in (V^\perp \cap H_1^2(M))/\{0\}} \frac{D(\phi,\phi)}{\int_M |\phi|^2}. 
\end{align*}
\\\\
References for the material in this appendix are: Canazani \cite[Chapter 6 \& 7]{canzani2013analysis}, Gilbarg and Trudinger \cite[Chapter 7]{gilbarg2015elliptic}, Schoen and Yau \cite[Chapter 3]{schoen1994lectures}. A complete survey of heat kernels on manifolds can be found in Grigoryan \cite{grigor2006heat}.

\chapter{Basic Spectral Graph Theory}
\label{AppendixE}
\section{Laplace Operator on a Graph}
Suppose $G=(E,V,w)$ is a simple weighted graph with a finite number of nodes and edges. The degree $w(u)$ is defined as 
\begin{align*}
          w(u):=\sum_{v} w(u,v),
\end{align*} 
The Laplacian $\mathcal{L}$ is given by
\begin{align*}
\mathcal{L}(u,v) =\begin{cases}
                             w(u) &  \text{ if } u=v\\
                             -w(u,v)      &  \text{ if $u$ and $v$ are adjacent}\\
                             0               &  \text{ otherwise}
                             \end{cases}
\end{align*}
Normalize the Laplacian $\mathcal{L}$ by degree matrix $D$
\begin{align*}
 \bar{\mathcal{L}} = D^{-\frac{1}{2}}\cdot \mathcal{L}\cdot D^{-\frac{1}{2}}.
\end{align*} 
Then we get 
\begin{align*}
\bar{\mathcal{L}}(u,v) =\begin{cases}
                             1 &  \text{ if } u=v\\
                             -\frac{w(u,v)}{\sqrt{w(u)\cdot w(v)}}      &  \text{ if $u$ and $v$ are adjacent}\\
                             0               &  \text{ otherwise}
                             \end{cases}
\end{align*}
\subsubsection{Note}
\begin{itemize}
\item The simple graph has no loop on any vertex, i.e. $w(u,u)=0$ for any $u \in V$.
\item The weight function $w:V\times V \rightarrow \mathbb{R}$ satisfies 
         \begin{align*}
                w(u,v) &= w(v,u),\\
                w(u,v)&\geq 0.
         \end{align*}
         In particular, $w(u,v)=0$ if $\{u,v\} \notin E$.
\end{itemize}
The real-valued function $f: V\rightarrow \mathbb{R}$ can be regarded as a $|V|$-dimensional vector $f:=(f(u))_{u\in V}$. Then the linear operator $\mathcal{L}$ on a graph is defined by 
\begin{align*}
   [\mathcal{L}f](u):&=\sum_{v:u\sim v} (f(u)-f(v))\cdot w(u,v) \\
                          &=f(u)-\frac{1}{w(u)}\sum_{v:u\sim v} f(v) w(u,v).
\end{align*}
\subsubsection{Note}
\begin{itemize}
\item In some literature, the Laplace operator has an opposite sign, i.e. 
\begin{align*}
[\mathcal{L}f](u):= -\sum_{v:u\sim v} (f(u)-f(v))\cdot w(u,v).
\end{align*}
\item On a connected graph $G$ with $|V|>1$, all the eigenvalues of $\mathcal{L}$ are in the interval $[0,2]$. In particular, eigenvalue $\lambda_1$ (=0) has algebraic multiplicity one.
\end{itemize}
Suppose $f$ and $g$ are real-valued functions on a graph. The inner product of them is defined by 
\begin{align} 
    (f,g)_G:=\sum_{u\in V} f(u)g(u) \cdot w(u).
\end{align}
The Rayleigh quotient of $\mathcal{L}$ is given by
\begin{align*}
       \mathcal{R}(f):&=\frac{(\mathcal{L}f,f)_G}{(f,f)_G} \\
                              &=\frac{\frac{1}{2}\sum_{u,v \in V}(f(u)-f(v))^2\cdot w(u,v)}{\sum_{u \in V}f(u)^2\cdot w(u)}\\
                              &=\frac{\sum_{u\sim v}(f(u)-f(v))^2\cdot w(u,v)}{\sum_{u \in V}f(u)^2\cdot w(u)}.
\end{align*}
According to the Courant-Fisher principle and the inner product on a graph (E.1.1), we have
\begin{align*}
      \lambda_k &=\inf_{f \perp D^{\frac{1}{2}}(V_{k-1})}\frac{\sum_{u\sim v}(f(u)-f(v))^2\cdot w(u,v)}{\sum_{u \in V}f(u)^2\cdot w(u)}\\
             &=\sup_{f\perp D^{\frac{1}{2}}(V_{k}^\perp)}\frac{\sum_{u\sim v}(f(u)-f(v))^2\cdot w(u,v)}{\sum_{u \in V}f(u)^2\cdot w(u)},
\end{align*}
where $V_k$ is the linear subspace spanned by eigenvector $v_1,\dots, v_{k-1}$.
\section{Random Walks on a Graph}
A random walk on a graph is a sequence of random variables $\{X_i\}_{i=0}^\infty$ on the vertices of a graph $G=(E,V,w)$ following the transition probability
\begin{align*}
        P(X_{i+1}=v|X_i=u) = \begin{cases}
                                                  \frac{w(u,v)}{w(u)} & v\sim u \\
                                                  0                    & v\not\sim u
                                           \end{cases}
\end{align*}
The entry $P(u,v)$ of a $|V|\times|V|$ matrix $P$ is defined by 
\begin{align*}
          P(u,v):=P(X_{i+1}=v|X_i=u).
\end{align*}
Then $P$ is a Markov kernel, i.e., for each vertex $u$
\begin{align*}
        \sum_{v:v\sim u} P(u,v) =1.
\end{align*}
By the Markov property \cite{karatzas2012brownian}
\begin{align*}
    P(X_{n+1}=v_{n+1},\dots, X_1=v_1|X_0=u) = P(v_{n+1},v_{n})\times\dots\times P(v_1,u),
\end{align*}
we have the $k$-th step transition kernel $P^k=P\times\dots\times P$ (k-fold).

\section{Heat equation on a graph}
Take the spectral decomposition of the normalized Laplacian matrix $\bar{\mathcal{L}}$
\begin{align*}
    \bar{\mathcal{L}}=U \cdot \Lambda \cdot U^T,
\end{align*}
where the columns of $U$ are $\phi_1,\dots, \phi_{|V|}$ and $\Lambda$ is a diagonal matrix with entries $\lambda_1, \dots, \lambda_{|V|}$.\\\\
The heat equation on a graph is defined by
\begin{align*}
      \frac{\partial u_t}{\partial t}=-\bar{\mathcal{L}}u_t,
\end{align*}
where $u_t$ is any real-valued function defined on a graph with respect to time $t$. \\\\
The heat kernel $H_t$ on a graph is defined as a $|V|\times |V|$ matrix such that for any time $t\geq0$, 
\begin{align*}
          H_t:=e^{-t\bar{\mathcal{L}}}.
\end{align*}
The heat kernel $H_t$ can be computed by the Laplacian eigenspectrum, i.e. 
\begin{align*}
      H_t=\sum_{i=1}^{|V|} e^{-\lambda_it}\phi_i\phi_i^T= U \cdot e^{-\Lambda t}\cdot U^T.
\end{align*}
In particular, 
\begin{align*}
H_0=I.
\end{align*}
References for the material in this appendix are: Chuang \cite[Chapter 1 \& 10]{chung1997spectral} and Grigoryan \cite[Chapter 2]{grigoryan2009analysis}. 


\bibliographystyle{alpha}
\cleardoublepage 
\phantomsection  
\renewcommand*{\bibname}{References}
\addcontentsline{toc}{chapter}{\textbf{References}}
\bibliography{uw-ethesis}

\begin{thebibliography}{BDSLT00}

\bibitem[AMS09]{absil2009optimization}
P-A Absil, Robert Mahony, and Rodolphe Sepulchre.
\newblock {\em Optimization algorithms on matrix manifolds}.
\newblock Princeton University Press, 2009.

\bibitem[Ban15]{BAsTEN}
Afonso~S. Banderia.
\newblock Ten lectures and forty-two open problems in the mathematics of data
  science.
\newblock {\em Lecture Notes}, 2015.

\bibitem[BDSLT00]{bernstein2000graph}
Mira Bernstein, Vin De~Silva, John~C Langford, and Joshua~B Tenenbaum.
\newblock Graph approximations to geodesics on embedded manifolds.
\newblock Technical report, Technical report, Department of Psychology,
  Stanford University, 2000.

\bibitem[Bel03]{belkin2003problems}
Mikhail Belkin.
\newblock Problems of learning on manifolds.
\newblock 2003.

\bibitem[BN03]{belkin2003laplacian}
Mikhail Belkin and Partha Niyogi.
\newblock Laplacian eigenmaps for dimensionality reduction and data
  representation.
\newblock {\em Neural computation}, 15(6):1373--1396, 2003.

\bibitem[Boc41]{bochner1941hilbert}
Salomon Bochner.
\newblock Hilbert distances and positive definite functions.
\newblock {\em Annals of Mathematics}, pages 647--656, 1941.

\bibitem[Can13]{canzani2013analysis}
Yaiza Canzani.
\newblock Analysis on manifolds via the laplacian.
\newblock {\em Lecture Notes, Harvard University, http://www. math. harvard.
  edu/~canzani/math253. html}, 2013.

\bibitem[CC00]{cox2000multidimensional}
Trevor~F Cox and Michael~AA Cox.
\newblock {\em Multidimensional scaling}.
\newblock CRC press, 2000.

\bibitem[Che50]{chern1950lecture}
Shiing-Shen Chern.
\newblock Lecture note on differential geometry.
\newblock {\em Chicago Univ}, 1950.

\bibitem[Chu97]{chung1997spectral}
Fan~RK Chung.
\newblock {\em Spectral graph theory}, volume~92.
\newblock American Mathematical Soc., 1997.

\bibitem[CL06]{coifman2006diffusion}
Ronald~R Coifman and St{\'e}phane Lafon.
\newblock Diffusion maps.
\newblock {\em Applied and computational harmonic analysis}, 21(1):5--30, 2006.

\bibitem[DG03]{donoho2003hessian}
David~L Donoho and Carrie Grimes.
\newblock Hessian eigenmaps: Locally linear embedding techniques for
  high-dimensional data.
\newblock {\em Proceedings of the National Academy of Sciences},
  100(10):5591--5596, 2003.

\bibitem[EAS98]{edelman1998geometry}
Alan Edelman, Tom{\'a}s~A Arias, and Steven~T Smith.
\newblock The geometry of algorithms with orthogonality constraints.
\newblock {\em SIAM journal on Matrix Analysis and Applications},
  20(2):303--353, 1998.

\bibitem[Fan50]{fan1950theorem}
Ky~Fan.
\newblock On a theorem of weyl concerning eigenvalues of linear transformations
  ii.
\newblock {\em Proceedings of the National Academy of Sciences}, 36(1):31--35,
  1950.

\bibitem[Gri06]{grigor2006heat}
Alexander Grigoryan.
\newblock Heat kernels on weighted manifolds and applications.
\newblock {\em Cont. Math}, 398:93--191, 2006.

\bibitem[Gri09]{grigoryan2009analysis}
Alexander Grigoryan.
\newblock Analysis on graphs.
\newblock {\em Lecture Notes, University Bielefeld}, 2009.

\bibitem[GT15]{gilbarg2015elliptic}
David Gilbarg and Neil~S Trudinger.
\newblock {\em Elliptic partial differential equations of second order}.
\newblock springer, 2015.

\bibitem[HAVL05]{hein2005graphs}
Matthias Hein, Jean-Yves Audibert, and Ulrike Von~Luxburg.
\newblock From graphs to manifolds--weak and strong pointwise consistency of
  graph laplacians.
\newblock In {\em International Conference on Computational Learning Theory},
  pages 470--485. Springer, 2005.

\bibitem[Jos08]{jost2008riemannian}
J{\"u}rgen Jost.
\newblock {\em Riemannian geometry and geometric analysis}.
\newblock Springer Science \& Business Media, 2008.

\bibitem[KS12]{karatzas2012brownian}
Ioannis Karatzas and Steven Shreve.
\newblock {\em Brownian motion and stochastic calculus}, volume 113.
\newblock Springer Science \& Business Media, 2012.

\bibitem[Laf04]{lafon2004diffusion}
St{\'e}phane~S Lafon.
\newblock {\em Diffusion maps and geometric harmonics}.
\newblock PhD thesis, Yale University, 2004.

\bibitem[Nas54]{nash1954c1}
John Nash.
\newblock C1 isometric imbeddings.
\newblock {\em Annals of mathematics}, pages 383--396, 1954.

\bibitem[Nas56]{nash1956imbedding}
John Nash.
\newblock The imbedding problem for riemannian manifolds.
\newblock {\em Annals of mathematics}, pages 20--63, 1956.

\bibitem[Pea01]{peason1901lines}
K~Peason.
\newblock On lines and planes of closest fit to systems of point in space.
\newblock {\em Philosophical Magazine}, 2:559--572, 1901.

\bibitem[Pet06]{petersen2006riemannian}
Peter Petersen.
\newblock {\em Riemannian geometry}, volume 171.
\newblock Springer, 2006.

\bibitem[RS78]{reed1978iv}
Michael Reed and Barry Simon.
\newblock {\em IV: Analysis of Operators}, volume~4.
\newblock Elsevier, 1978.

\bibitem[RS00]{roweis2000nonlinear}
Sam~T Roweis and Lawrence~K Saul.
\newblock Nonlinear dimensionality reduction by locally linear embedding.
\newblock {\em Science}, 290(5500):2323--2326, 2000.

\bibitem[SC10]{saloff2010heat}
Laurent Saloff-Coste.
\newblock The heat kernel and its estimates.
\newblock {\em Probabilistic approach to geometry}, 57:405--436, 2010.

\bibitem[Sun12]{SunMaxA}
Y.-K. Sun.
\newblock Matrix analysis notes.
\newblock {\em Lecture Notes}, 2012.

\bibitem[SW12]{singer2012vector}
Amit Singer and H-T Wu.
\newblock Vector diffusion maps and the connection laplacian.
\newblock {\em Communications on pure and applied mathematics},
  65(8):1067--1144, 2012.

\bibitem[SWW00]{smolyanov2000brownian}
OG~Smolyanov, HV~Weizs{\"a}cker, and O~Wittich.
\newblock Brownian motion on a manifold as limit of stepwise conditioned
  standard brownian motions.
\newblock {\em Stochastic processes, physics and geometry: new interplays, II},
  29:589--602, 2000.

\bibitem[SY94]{schoen1994lectures}
Richard Schoen and Shing-Tung Yau.
\newblock {\em Lectures on differential geometry}.
\newblock 1994.

\bibitem[TDSL00]{tenenbaum2000global}
Joshua~B Tenenbaum, Vin De~Silva, and John~C Langford.
\newblock A global geometric framework for nonlinear dimensionality reduction.
\newblock {\em science}, 290(5500):2319--2323, 2000.

\bibitem[Tor58]{torgerson1958theory}
Warren~S Torgerson.
\newblock Theory and methods of scaling.
\newblock 1958.

\bibitem[VLBB08]{von2008consistency}
Ulrike Von~Luxburg, Mikhail Belkin, and Olivier Bousquet.
\newblock Consistency of spectral clustering.
\newblock {\em The Annals of Statistics}, pages 555--586, 2008.

\bibitem[Whi34]{whitney1934analytic}
Hassler Whitney.
\newblock Analytic extensions of differentiable functions defined in closed
  sets.
\newblock {\em Transactions of the American Mathematical Society},
  36(1):63--89, 1934.

\bibitem[Wit05]{wittman}
Todd Wittman.
\newblock {\em MANI fold learning MATLAB demo}.
\newblock 2005.

\bibitem[WS06]{weinberger2006unsupervised}
Kilian~Q Weinberger and Lawrence~K Saul.
\newblock Unsupervised learning of image manifolds by semidefinite programming.
\newblock {\em International Journal of Computer Vision}, 70(1):77--90, 2006.

\bibitem[YH38]{young1938discussion}
Gale Young and Alston~S Householder.
\newblock Discussion of a set of points in terms of their mutual distances.
\newblock {\em Psychometrika}, 3(1):19--22, 1938.

\bibitem[ZZ04]{zhang2004principal}
Zhen-yue Zhang and Hong-yuan Zha.
\newblock Principal manifolds and nonlinear dimensionality reduction via
  tangent space alignment.
\newblock {\em Journal of Shanghai University (English Edition)},
  8(4):406--424, 2004.

\end{thebibliography}


\nocite{}
\end{document}